\documentclass[10pt,twocolumn,letterpaper]{article}

\usepackage[pagenumbers]{cvpr} 

\definecolor{cvprblue}{rgb}{0.21,0.49,0.74}
\usepackage[pagebackref,breaklinks,colorlinks,allcolors=cvprblue]{hyperref}

\usepackage{cite}
\usepackage{amsmath,amssymb}
\usepackage[ruled]{algorithm2e}
\usepackage{textcomp}
\usepackage{multirow}
\usepackage{algorithmic}
\usepackage{colortbl}
\usepackage{graphicx} 
\usepackage{adjustbox}
\usepackage[table]{xcolor}
\usepackage{subcaption}
\usepackage{wrapfig}
\usepackage{enumitem}
\usepackage{amsthm}

\definecolor{lightgray}{gray}{0.9}
\definecolor{lightblue}{rgb}{0.68, 0.85, 0.9}
\makeatletter

\DeclareRobustCommand\onedot{\futurelet\@let@token\@onedot}
\def\@onedot{\ifx\@let@token.\else.\null\fi\xspace}
\def\eg{\emph{e.g}\onedot}

\makeatother

\def\paperID{} 
\def\confName{CVPR}
\def\confYear{2026}

\title{MultiModal Fine-tuning with Synthetic Captions}

\author{Shohei Enomoto\\
NTT\\
Tokyo, Japan\\
{\tt\small shohei.enomoto@ntt.com}
\and
Shin’ya Yamaguchi\\
NTT\\
Tokyo, Japan\\
{\tt\small shinya.yamaguchi@ntt.com}
}

\begin{document}
\maketitle

\begin{abstract}
In this paper, we address a fundamental gap between pre-training and fine-tuning of deep neural networks: while pre-training has shifted from unimodal to multimodal learning with enhanced visual understanding, fine-tuning predominantly remains unimodal, limiting the benefits of rich pre-trained representations.
To bridge this gap, we propose a novel approach that transforms unimodal datasets into multimodal ones using Multimodal Large Language Models (MLLMs) to generate synthetic image captions for fine-tuning models with a multimodal objective.
Our method employs carefully designed prompts incorporating class labels and domain context to produce high-quality captions tailored for classification tasks.
Furthermore, we introduce a supervised contrastive loss function that explicitly encourages clustering of same-class representations during fine-tuning, along with a new inference technique that leverages class-averaged text embeddings from multiple synthetic captions per image.
Extensive experiments across 13 image classification benchmarks demonstrate that our approach outperforms baseline methods, with particularly significant improvements in few-shot learning scenarios.
Our work establishes a new paradigm for dataset enhancement that effectively bridges the gap between multimodal pre-training and fine-tuning.
Our code is available at \url{https://github.com/s-enmt/MMFT}.

\end{abstract}

\section{Introduction}
\label{sec:intro}

Fine-tuning based on pre-trained weights is the de facto standard approach for training deep neural networks on a new task~\citep{he2019rethinking}.
In computer vision, pre-training methodologies have undergone a paradigm shift from unimodal learning, which employs datasets consisting of images and categorical labels trained with cross-entropy loss~\citep{russakovsky2015imagenet}, to multimodal learning, which leverages image-caption pairs through contrastive learning~\citep{clip,align}.
This evolution has led to state-of-the-art models with enhanced visual understanding and representation capabilities.

However, a significant gap between pre-training and fine-tuning has emerged: despite the advances in pre-training, fine-tuning still predominantly relies on legacy unimodal datasets with categorical labels~\citep{mmrl,clip_ast,rada}.
This disparity between multimodal pre-training and unimodal fine-tuning creates a fundamental mismatch, which is amplified under few-shot fine-tuning because the model readily overfits to the task-specific distribution, preventing downstream tasks from fully benefiting from the rich representations learned during multimodal pre-training~\citep{mmrl}.
This discrepancy can be partially attributed to the structural limitations of existing datasets, which lack the textual information to fully exploit multimodal learning advantages.

To address this challenge, we propose an approach for transforming unimodal datasets into multimodal ones by using Multimodal Large Language Models (MLLMs)~\citep{instructblip,blip2,gemini,openai2025gpt5,gpt4o}.
Our method uses class labels, domain context, and specific visual characteristics in carefully designed prompts to guide MLLMs in generating high-quality image captions tailored for image classification tasks.
Furthermore, we introduce novel fine-tuning and inference techniques specifically designed for our synthesized multimodal datasets.
Our novel fine-tuning approach incorporates a supervised contrastive loss function that uses class label information to guide the alignment of semantically similar images in the embedding space.
Unlike standard contrastive learning, which might map images of the same class to different regions in the representation space, our method explicitly encourages clustering of same-class representations.
Additionally, while conventional approaches like CLIP~\citep{clip} typically use a single text template per class (\eg, ``a photo of a \texttt{[class\_name]}''), our synthetic multimodal dataset provides multiple unique captions per image, offering richer and more specific information.
To use this enhanced data structure, we propose a novel inference method that averages text embeddings for each class and computes similarity between image embeddings and these averaged class-specific text embeddings.
Through these techniques, we establish a new paradigm for dataset enhancement that effectively bridges the gap between multimodal pre-training and fine-tuning, improving performance on downstream tasks.

Extensive experiments across various image classification benchmarks demonstrate that our approach outperforms baseline methods.
Notably, in few-shot learning scenarios, our method shows substantial improvements over existing techniques.
Furthermore, our generated captions enable zero-training image classification that surpasses fine-tuned approaches with 1, 4, and 8 shots per class, demonstrating the high discriminative power of our synthetic captions, particularly in data-constrained scenarios.

Our main contributions are as follows:
\begin{itemize}[leftmargin=*]
\setlength{\itemsep}{2pt}
\item We propose an approach that transforms unimodal downstream task datasets into multimodal ones using MLLMs, bringing the benefits of multimodality to the fine-tuning stage and bridging the gap between multimodal pre-training and task-specific adaptation.
\item We propose a novel fine-tuning method using supervised contrastive loss and a new inference approach that leverages all generated synthetic captions with rich information, both specifically designed for our synthetic multimodal datasets.
\item We conducted comprehensive experiments across various benchmarks. In particular, in scenarios with limited training data, our approach shows highly impressive results.
\end{itemize}

\section{Related Work}
\label{sec:related}


\paragraph{CLIP Fine-tuning}
Traditional fine-tuning approaches for CLIP-like models~\citep{clip,openclip,siglip,align,florence,sic,slip,flip} include linear probing and full fine-tuning, which effectively improve in-distribution performance but often compromise out-of-distribution robustness~\citep{lp_ft,wise_ft}.
Several strategies have been proposed to address this challenge, including weight averaging~\citep{wise_ft,wortsman2022model,kumar2022calibrated} and prompt learning methods~\citep{zhou2022coop, zhou2022cocoop,zheng2024large}.
Recent work has also explored parameter-efficient fine-tuning through adapter-based methods~\citep{clip_adapter,mma}, learnable representation spaces~\citep{mmrl}, adaptive parameter selection~\citep{clip_ast}, and rational matrix adaptation~\citep{rada}.
Most relevant to our work, FLYP~\citep{flyp} fine-tunes both encoders using contrastive learning with a class-based text template.

Our approach differs from previous work by focusing on enhancing fine-tuning through MLLM-generated captions and supervised contrastive learning.
We transform existing unimodal datasets into multimodal ones and leverage the rich information in generated captions to improve classification performance, particularly in few-shot scenarios.

\paragraph{Synthetic Caption Generation}
Recent studies have explored synthetic caption generated by large models to improve the classification performance of CLIP.
There are approaches using LLM-generated captions to enhance zero-shot classification capabilities~\citep{dclip,waffle,saha2024improved,comparative}, text rewriting for data augmentation~\citep{fan2023improving,lai2024veclip}, and research utilizing fully synthetic datasets for pre-training~\citep{hammoud2024synthclip}.
More recently, LatteCLIP~\citep{cao2025latteclip} leveraged captions generated by MLLM for unsupervised learning of CLIP.

Unlike previous works that primarily focus on zero-shot classification, pre-training, or unsupervised learning, our research emphasizes fine-tuning on downstream tasks.
We propose a novel method that brings the benefits of multimodality to downstream tasks through MLLM-generated captions, supervised contrastive fine-tuning, and class-averaged inference.

\section{Preliminaries}
\label{sec:preliminaries}

CLIP (Contrastive Language-Image Pre-training)~\citep{clip} consists of an image encoder $f_I$ and a text encoder $f_T$, which map inputs to $d$-dimensional embeddings $\mathbf{v}_I = f_I(I)$ and $\mathbf{v}_T = f_T(T)$, respectively.
The model is pre-trained on image-text pairs using contrastive learning, which aligns matching pairs in a shared embedding space.

Given a batch of $N$ image-text pairs $\{(I_i, T_i)\}_{i=1}^N$, CLIP computes a similarity matrix $S$ where $S_{ij} = \frac{\mathbf{v}_{I_i} \cdot \mathbf{v}_{T_j}}{||\mathbf{v}_{I_i}|| \cdot ||\mathbf{v}_{T_j}||}$.
The contrastive loss is then computed as:

\begin{equation}
\mathcal{L}_I = -\frac{1}{N}\sum_{i=1}^N \log \frac{\exp(S_{ii}/\tau)}{\sum_{j=1}^N \exp(S_{ij}/\tau)},
\end{equation}

\begin{equation}
\mathcal{L}_T = -\frac{1}{N}\sum_{i=1}^N \log \frac{\exp(S_{ii}/\tau)}{\sum_{j=1}^N \exp(S_{ji}/\tau)},
\end{equation}

\begin{equation}
\label{eq:l_std}
\mathcal{L}^{std} = \frac{\mathcal{L}_I + \mathcal{L}_T}{2},
\end{equation}
where $\tau$ is a temperature parameter.

For zero-shot classification, CLIP: (1) constructs text templates like ``a photo of a \texttt{[class\_name]}'' for each class, (2) computes text embeddings for these templates, (3) computes the image embedding, (4) calculates cosine similarities between the image and each class embedding, and (5) predicts the class with highest similarity.
This enables classification without task-specific fine-tuning, leveraging the alignment between visual and textual representations.

\section{Proposed Method}
\label{sec:proposed}

\begin{figure*}[t]
\centering
\includegraphics[width=0.8\linewidth]{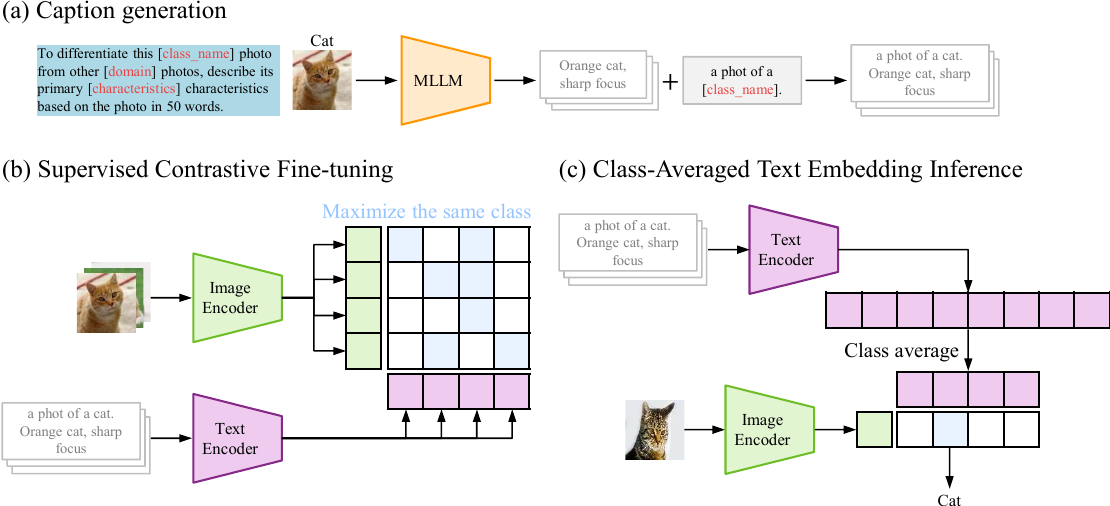}
\caption{
Overview of our proposed approach. 
Our method consists of three key components: (a) Synthetic multimodal dataset generation using MLLM, (b) Supervised contrastive fine-tuning that leverages both image-caption pairs and class information, and (c) Class-averaged text embedding inference.
}
\label{fig:method_overview}
\end{figure*}

In this section, we present our approach for bridging the gap between multimodal pre-training and unimodal fine-tuning through dataset transformation.
Our method consists of three main components: (a) synthetic caption generation from unimodal datasets with MLLM, (b) supervised contrastive fine-tuning that leverages both image-caption pairs and class information, and (c) a novel inference method based on class-averaged text embeddings.
\Cref{fig:method_overview} illustrates the overall framework of our proposed approach.

\subsection{Synthetic Multimodal Dataset Generation}

Given a unimodal dataset consisting of image-label pairs $(I, Y)$, our goal is to generate synthetic captions that effectively describe the visual characteristics of each image while maintaining semantic consistency with its class label.
The key innovation in our approach is the design of prompts that strategically incorporate three essential elements: class name information, dataset domain context, and salient visual features to guide the MLLM's caption generation process. 
Our prompt template is structured as:
``To differentiate this \texttt{[class\_name]} photo from other \texttt{[domain]} photos, describe its primary \texttt{[characteristics]} characteristics based on the photo in 50 words.''
The template parameters include \texttt{[class\_name]}, which is the category label of the image (e.g., ``daisy'' for the Flowers102 dataset), and \texttt{[domain]} representing the dataset context (e.g., ``flowers'' for the Flowers102 dataset). 
We generate three distinct captions per image by using three different \texttt{[characteristics]}: visual, shape, and texture. 
The visual characteristic provides the most intuitive description, while shape and texture are based on previous research showing that DNN models recognize objects primarily through these two properties~\citep{shape_texture,vit_shape_texture}.
We fix the caption length to 50 words to ensure the tokenized caption fits within CLIP's maximum context length of 77 tokens while providing sufficient descriptive information.

After generating the initial caption using the MLLM, we prepend ``a photo of a \texttt{[class\_name]}. '' to create the final caption for each image. 
This standard CLIP-style prefix ensures class information is explicitly present, maintains compatibility with CLIP's pre-training format, and provides a consistent reference point, while the MLLM-generated portion adds rich discriminative details.

The resulting synthetic multimodal dataset consists of triplets $(I, Y, \mathbb{C})$, where each image is paired with both its original class label and multiple generated captions.
This structure enables more effective fine-tuning by leveraging both the rich visual descriptions from captions and the categorical information from labels.

\subsection{Supervised Contrastive Fine-tuning with Synthetic Multimodal Datasets}

We propose a novel fine-tuning approach that leverages our synthetic multimodal dataset triplets $(I, Y, \mathbb{C})$ to more effectively transfer knowledge from pre-trained CLIP models to downstream tasks.
Conventional contrastive learning with image-caption pairs, as described in \Cref{sec:preliminaries}, does not explicitly use class label information during training.
This can lead to suboptimal embeddings where semantically similar images (i.e., those from the same class) may be mapped to different regions in the embedding space, limiting the model's discriminative capabilities.

Our approach builds upon two important research directions in contrastive learning: UniCL~\citep{yang2022unified} and SupCon~\citep{supcon}.
While these methods have shown promise, they either focus primarily on the pre-training stage or don't fully exploit the rich information available in synthetic captions for downstream tasks.
Our approach extends these ideas to fine-tuning with synthetic multimodal datasets, where we have access to both rich captions and class labels for each image.

To address the limitations of conventional contrastive learning, we introduce a supervised contrastive loss function that explicitly incorporates class label information to guide the alignment of semantically similar images and captions in the embedding space.
Our approach extends the standard contrastive loss defined in \Cref{eq:l_std} by adding class supervision signals.

Our supervised contrastive loss incorporates class information through a similarity mask $M \in \mathbb{R}^{N \times N}$:

\begin{equation}
M_{ij} = 
\begin{cases}
1, & \text{if } Y_i = Y_j \\
0, & \text{otherwise}
\end{cases}
\end{equation}

After excluding self-pairs with $\hat{M} = M - I$, we compute our supervised loss as:

\begin{equation}
\mathcal{L}^{sup} = -\frac{1}{|\mathcal{V}|}\sum_{i \in \mathcal{V}} \frac{\sum_{j=1}^{N} \hat{M}_{ij} \cdot \log \frac{\exp(S_{ij})}{\sum_{k=1}^{N} \exp(S_{ik})}} {\sum_{j=1}^{N} \hat{M}_{ij}} ,
\end{equation}
where $\mathcal{V} = \{i | \sum_{j=1}^{N} \hat{M}_{ij} > 0\}$ is the set of samples with at least one positive pair.
Our final loss function combines this with the standard CLIP loss:

\begin{equation}
\mathcal{L} = (1-w) \cdot \mathcal{L}^{std} + w \cdot \mathcal{L}^{sup},
\end{equation}
where $w$ is the hyperparameter that weights the loss.

For multiple captions per image, we implement a data augmentation strategy by randomly selecting one caption per image during each training iteration, effectively increasing training diversity without increasing batch size.

\subsection{Class-Averaged Text Embedding Inference}

While standard CLIP inference creates a single text template for each class (e.g., ``a photo of a \texttt{[class\_name]}'') and measures similarity between image embeddings and these template embeddings, this approach fails to fully leverage the rich and diverse information contained in our synthetic multimodal datasets.
Since our synthetic multimodal dataset provides multiple captions per image, we propose a novel inference method that effectively utilizes this enhanced information to improve classification accuracy.

Our key insight is that the diverse captions generated for images of the same class collectively provide a more comprehensive representation of the visual characteristics that define that class.
By aggregating these caption embeddings, we can create a more robust class representation that captures the essential features that distinguish each class.

For each class $k \in \{1,2,...,K\}$, we collect all associated captions $\mathbb{C}_k = \{C_{k1}, C_{k2}, ..., C_{kn_k}\}$, where $n_k$ is the number of captions for class $k$, and encode them using the text encoder.
We then compute the class-averaged text embedding by first normalizing each caption embedding and then averaging them: $\mathbf{v}_{T_k} = \frac{1}{n_k} \sum_{i=1}^{n_k} \frac{\mathbf{v}_{C_{ki}}}{||\mathbf{v}_{C_{ki}}||}$, where $\mathbf{v}_{C_{ki}}$ is the text embedding of caption $C_{ki}$.
The final class embedding is further normalized: $\hat{\mathbf{v}}_{T_k} = \frac{\mathbf{v}_{T_k}}{||\mathbf{v}_{T_k}||}$.
During inference, we encode the query image $I$ to obtain the image embedding, then compute its cosine similarity with each pre-computed class-averaged text embedding.
The class with the highest similarity score is predicted as the output.
This approach captures the essential characteristics that define each class through multiple perspectives, resulting in more robust classification, particularly for few-shot scenarios.

\section{Experiments}
\label{sec:experiments}

To evaluate our approach, we conducted comprehensive experiments on various image classification tasks, including standard fine-tuning and few-shot learning scenarios. 

\subsection{Experimental Setup}

\subsubsection{Datasets}

We evaluated our approach on a diverse set of 13 image classification datasets.
For general object recognition, we used CIFAR10~\citep{cifar}, CIFAR100~\citep{cifar}, and Caltech101~\citep{caltech101}. 
Fine-grained classification tasks are represented by CUB-200-2011~\citep{cub}, Stanford Cars~\citep{cars}, Oxford-IIIT Pet~\citep{pets}, and Food-101~\citep{food}. 
We also included specialized domains such as DTD~\citep{dtd}, Oxford Flowers-102~\citep{flowers}, EuroSAT~\citep{eurosat}, GTSRB~\citep{gtsrb}, and FGVC-Aircraft~\citep{air}.
For few-shot settings, we additionally use ImageNet~\citep{imagenet}, which contains 1,000 object categories and serves as a comprehensive benchmark for testing generalization capabilities.
We used 12 datasets, except for ImageNet, for the fine-tuning setting, and all 13 datasets for the few-shot settings.
\Cref{tab:datasets} provides statistics for each dataset, including the number of classes, data split, and \texttt{[domain]} characteristics.

\begin{table}[tb!]
\centering
\small
\caption{
Statistics of datasets used in our experiments.
Pet, Flowers, and DTD use the torchvision~\citep{torchvision} train/val/test splits.
EuroSAT and Caltech101 use the splits provided by CoOp~\citep{zhou2022coop,zhou2022cocoop}.
For ImageNet, we use 10,000 samples from the training set as validation.
For all other datasets, we randomly split the training data into train/val with a 0.9:0.1 ratio.
}
\label{tab:datasets}
\resizebox{\linewidth}{!}{
\begin{tabular}{lccccc}
\toprule
Dataset & \#Classes & \#Train & \#Val & \#Test & Domain \\ \midrule
CIFAR10 & 10 & 45,000 & 5,000 & 10,000 & general objects \\
CIFAR100 & 100 & 45,000 & 5,000 & 10,000 & general objects \\
CUB & 200 & 5,394 & 600 & 5,794 & birds \\
Caltech101 & 100 & 4,128 & 1,649 & 2,465 & general objects \\
Cars & 196 & 7,329 & 815 & 8,041 & cars \\
DTD & 47 & 1,880 & 1,880 & 1,880 & textures \\
EuroSAT & 10 & 13,500 & 5,400 & 8,100 & satellite \\
Flowers & 102 & 1,020 & 1,020 & 6,149 & flowers \\
Food & 101 & 68,175 & 7,575 & 25,250 & food \\
GTSRB & 43 & 23,976 & 2,664 & 12,630 & traffic signs \\
Pet & 37 & 3,312 & 368 & 3,669 & pets \\
Air & 100 & 3,334 & 3,333 & 3,333 & aircrafts \\
ImageNet & 1,000 & 1,271,167 & 10,000 & 50,000 & general objects \\ 
\bottomrule
\end{tabular}
}
\end{table}

\subsubsection{Implementation Details}

We use two backbone architectures from the OpenCLIP~\citep{openclip} library: ResNet-50~\citep{resnet} and ViT-B/32~\citep{vit}, with pre-trained weights provided by OpenAI from the CLIP dataset containing 400M image-text pairs. 
For caption generation, we employ Gemini 2.5 Flash-Lite~\citep{gemini} as our default MLLM due to its strong multimodal understanding capabilities and efficient inference. 
The captions are generated with a temperature of 0.2 to maintain coherence while allowing some diversity in the outputs.
For fine-tuning experiments, we use the AdamW~\citep{adamw} optimizer with a learning rate of $1e-5$ and a weight decay of $1e-4$. 
We employ a cosine annealing learning rate schedule with a minimum learning rate of $0.0$. 
Training is performed with a batch size of $64$ for $50$ epochs on standard fine-tuning tasks and few-shot learning scenarios.
For few-shot learning, we randomly sample K examples per class (K = 1, 4, 8, 16, 32) from the training set while maintaining class balance. 
For datasets with insufficient examples for certain shot counts, we use all available examples.
We repeat each experiment three times with different random seeds and report the mean and standard error.
We report Top-1 accuracy on the test set for all experiments.

\subsubsection{Baselines}

We compare our approach against several baselines in both fine-tuning and few-shot learning scenarios.

\paragraph{Zero-shot Classification Baselines}
\textbf{ZS} performs zero-shot classification without any training examples. With CLIP, ZS uses text templates (e.g., ``a photo of a \texttt{[class\_name]}''). 
With MLLM, ZS inputs an image with a prompt requesting the model to select from the class names. 
Following~\citet{cao2025latteclip}, the MLLM prompt is: ``Select the most appropriate category for the image from the following options: \texttt{[class\_names]}. Write only the category name.''
Here, \texttt{[class\_names]} is replaced with the list of all class names in the dataset.
\textbf{DCLIP}~\citep{dclip} uses class descriptions generated by an LLM instead of simple text templates for zero-shot classification.
\textbf{WaffleCLIP}~\citep{waffle} replaces portions of LLM-generated class descriptions with random strings to improve robustness.
\textbf{Comparative}~\citep{comparative} employs LLM-generated class descriptions that specifically focus on distinguishing similar classes.

\paragraph{Fine-tuning and Few-shot Learning Baselines}
\textbf{FT} applies standard full fine-tuning of the image encoder using cross-entropy loss with class labels.
\textbf{FLYP}~\citep{flyp} fine-tunes both the image and text encoders using contrastive learning with text templates.
\textbf{Comparative+Filtering}~\citep{comparative} computes average image features for each class using a small set of training examples, then filters the LLM-generated comparative descriptions based on CLIP similarity scores between descriptions and average image features. 
This approach does not involve model training.

\subsection{Results on Standard Fine-tuning}

We evaluated our approach against baselines across 12 diverse image classification datasets.
\Cref{tab:main} shows the comparison of different methods on standard fine-tuning tasks.

\begin{table*}[tb!]
\small
\centering
\caption{
Comparison of classification accuracy ($\%$) across 12 datasets with different methods. 
For our method, the loss weight $w$ is tuned per dataset by exploring values from 0 to 1.0 in 0.1 increments. 
The best results for each backbone model are in \textbf{bold}.
}
\resizebox{\linewidth}{!}{
\begin{tabular}{llccccccccccccc}
\toprule
Model & Method & Air & CIFAR10 & CIFAR100 & CUB & Caltech101 & Cars & DTD & EuroSAT & Flowers & Food & GTSRB & Pet & Average  \\ \midrule
MLLM &  ZS & 53.11 & 91.56 & 70.09 & 65.84 & 96.92 & 86.90 & 67.93 & 59.83 & 84.68 & 91.85 & 77.38 & 92.56 & 78.22  \\
 \midrule
\multirow{4}{*}{RN50} & ZS & 12.75 & 69.87 & 40.98 & 42.13 & 83.81 & 50.99 & 39.04 & 24.96 & 56.25 & 76.67 & 25.46 & 80.57 & 50.29   \\
 & FT & 27.57 \fontsize{5pt}{5pt}\selectfont{$\pm$ 0.31} & 95.79 \fontsize{5pt}{5pt}\selectfont{$\pm$ 0.07} & 81.83 \fontsize{5pt}{5pt}\selectfont{$\pm$ 0.04} & 60.87 \fontsize{5pt}{5pt}\selectfont{$\pm$ 0.46} & 92.90 \fontsize{5pt}{5pt}\selectfont{$\pm$ 0.21} & 64.89 \fontsize{5pt}{5pt}\selectfont{$\pm$ 0.28} & 67.75 \fontsize{5pt}{5pt}\selectfont{$\pm$ 0.32} & 98.29 \fontsize{5pt}{5pt}\selectfont{$\pm$ 0.04} & 69.29 \fontsize{5pt}{5pt}\selectfont{$\pm$ 0.21} & 86.32 \fontsize{5pt}{5pt}\selectfont{$\pm$ 0.11} & 97.20 \fontsize{5pt}{5pt}\selectfont{$\pm$ 0.23} & 85.99 \fontsize{5pt}{5pt}\selectfont{$\pm$ 0.25} & 77.39  \\
 & FLYP & 38.91 \fontsize{5pt}{5pt}\selectfont{$\pm$ 0.59} & 95.93 \fontsize{5pt}{5pt}\selectfont{$\pm$ 0.07} & 82.18 \fontsize{5pt}{5pt}\selectfont{$\pm$ 0.20} & 71.52 \fontsize{5pt}{5pt}\selectfont{$\pm$ 0.22} & 93.41 \fontsize{5pt}{5pt}\selectfont{$\pm$ 0.10} & 74.92 \fontsize{5pt}{5pt}\selectfont{$\pm$ 0.43} & 68.05 \fontsize{5pt}{5pt}\selectfont{$\pm$ 0.54} & 98.47 \fontsize{5pt}{5pt}\selectfont{$\pm$ 0.03} & 81.92 \fontsize{5pt}{5pt}\selectfont{$\pm$ 0.20} & 86.93 \fontsize{5pt}{5pt}\selectfont{$\pm$ 0.02} & 97.77 \fontsize{5pt}{5pt}\selectfont{$\pm$ 0.07} & 87.83 \fontsize{5pt}{5pt}\selectfont{$\pm$ 0.32} & 81.49  \\
\rowcolor{lightgray} & Ours & \textbf{40.09} \fontsize{5pt}{5pt}\selectfont{$\pm$ 0.42} & \textbf{96.64} \fontsize{5pt}{5pt}\selectfont{$\pm$ 0.03} & \textbf{82.55} \fontsize{5pt}{5pt}\selectfont{$\pm$ 0.04} & \textbf{72.20} \fontsize{5pt}{5pt}\selectfont{$\pm$ 0.17} & \textbf{93.63} \fontsize{5pt}{5pt}\selectfont{$\pm$ 0.11} & \textbf{75.60} \fontsize{5pt}{5pt}\selectfont{$\pm$ 0.06} & \textbf{68.88} \fontsize{5pt}{5pt}\selectfont{$\pm$ 0.20} & \textbf{98.54} \fontsize{5pt}{5pt}\selectfont{$\pm$ 0.03} & \textbf{82.88} \fontsize{5pt}{5pt}\selectfont{$\pm$ 0.06} & \textbf{87.61} \fontsize{5pt}{5pt}\selectfont{$\pm$ 0.14} & \textbf{97.92} \fontsize{5pt}{5pt}\selectfont{$\pm$ 0.04} & \textbf{88.40} \fontsize{5pt}{5pt}\selectfont{$\pm$ 0.10} & \textbf{82.08}  \\
 \midrule
\multirow{4}{*}{ViT} & ZS & 13.68 & 86.19 & 62.67 & 47.38 & 89.53 & 52.48 & 40.27 & 34.53 & 56.42 & 77.66 & 23.77 & 78.28 & 55.24 \\
 & FT & 38.93 \fontsize{5pt}{5pt}\selectfont{$\pm$ 0.37} & 97.45 \fontsize{5pt}{5pt}\selectfont{$\pm$ 0.06} & 88.23 \fontsize{5pt}{5pt}\selectfont{$\pm$ 0.03} & 72.07 \fontsize{5pt}{5pt}\selectfont{$\pm$ 0.27} & 96.11 \fontsize{5pt}{5pt}\selectfont{$\pm$ 0.11} & 77.63 \fontsize{5pt}{5pt}\selectfont{$\pm$ 0.17} & 74.72 \fontsize{5pt}{5pt}\selectfont{$\pm$ 0.17} & 98.82 \fontsize{5pt}{5pt}\selectfont{$\pm$ 0.01} & 90.25 \fontsize{5pt}{5pt}\selectfont{$\pm$ 0.13} & 87.23 \fontsize{5pt}{5pt}\selectfont{$\pm$ 0.07} & 98.93 \fontsize{5pt}{5pt}\selectfont{$\pm$ 0.17} & 92.21 \fontsize{5pt}{5pt}\selectfont{$\pm$ 0.26} & 84.38 \\
 & FLYP & \textbf{46.28} \fontsize{5pt}{5pt}\selectfont{$\pm$ 0.42} & 97.64 \fontsize{5pt}{5pt}\selectfont{$\pm$ 0.02} & 88.56 \fontsize{5pt}{5pt}\selectfont{$\pm$ 0.15} & \textbf{79.25} \fontsize{5pt}{5pt}\selectfont{$\pm$ 0.19} & 96.69 \fontsize{5pt}{5pt}\selectfont{$\pm$ 0.11} & \textbf{83.19} \fontsize{5pt}{5pt}\selectfont{$\pm$ 0.11} & 76.74 \fontsize{5pt}{5pt}\selectfont{$\pm$ 0.31} & 98.72 \fontsize{5pt}{5pt}\selectfont{$\pm$ 0.06} & 93.77 \fontsize{5pt}{5pt}\selectfont{$\pm$ 0.18} & 87.88 \fontsize{5pt}{5pt}\selectfont{$\pm$ 0.05} & 98.83 \fontsize{5pt}{5pt}\selectfont{$\pm$ 0.14} & 92.26 \fontsize{5pt}{5pt}\selectfont{$\pm$ 0.18} & 86.65  \\
\rowcolor{lightgray} & Ours & 45.65 \fontsize{5pt}{5pt}\selectfont{$\pm$ 0.64} & \textbf{98.04} \fontsize{5pt}{5pt}\selectfont{$\pm$ 0.04} & \textbf{88.92} \fontsize{5pt}{5pt}\selectfont{$\pm$ 0.08} & 78.75 \fontsize{5pt}{5pt}\selectfont{$\pm$ 0.16} & \textbf{96.73} \fontsize{5pt}{5pt}\selectfont{$\pm$ 0.07} & 82.25 \fontsize{5pt}{5pt}\selectfont{$\pm$ 0.33} & \textbf{77.68} \fontsize{5pt}{5pt}\selectfont{$\pm$ 0.10} & \textbf{98.97} \fontsize{5pt}{5pt}\selectfont{$\pm$ 0.03} & \textbf{94.09} \fontsize{5pt}{5pt}\selectfont{$\pm$ 0.05} & \textbf{88.79} \fontsize{5pt}{5pt}\selectfont{$\pm$ 0.03} & \textbf{99.08} \fontsize{5pt}{5pt}\selectfont{$\pm$ 0.02} & \textbf{92.77} \fontsize{5pt}{5pt}\selectfont{$\pm$ 0.10} & \textbf{86.81} \\ 
 \bottomrule
\end{tabular}
}
\label{tab:main}
\end{table*}

When using ResNet-50 as the backbone, Ours consistently outperforms all baseline methods across all 12 datasets.
Compared to FT, Ours shows substantial improvements, with an average accuracy gain of 4.69 percentage points ($82.08\%$ vs. $77.39\%$).
Even when compared to FLYP, which also uses contrastive learning but with generic text templates, our method achieves better performance with an average improvement of 0.59 percentage points ($82.08\%$ vs. $81.49\%$).
On the Aircraft dataset, our method achieves $40.09\%$ accuracy, outperforming FLYP by 1.18 percentage points.
For the ViT backbone, Ours outperforms the baselines on the majority of datasets, achieving the highest average accuracy of $86.81\%$.
Compared to FT, Ours shows a substantial improvement of 2.43 percentage points ($86.81\%$ vs. $84.38\%$).
When compared to FLYP, our method achieves a marginal improvement in average accuracy ($86.81\%$ vs. $86.65\%$), though the performance varies across individual datasets.
Compared to CLIP, MLLM shows significantly higher zero-shot classification performance.
Our approach effectively transfers this strong recognition capability from MLLM to smaller models through synthetic multimodal datasets, improving classification performance across diverse domains.
These results validate the effectiveness of our synthetic caption generation and supervised contrastive fine-tuning strategy.

\subsection{Few-shot Learning Results}

\begin{figure}[t]
\centering
\includegraphics[width=0.9\linewidth]{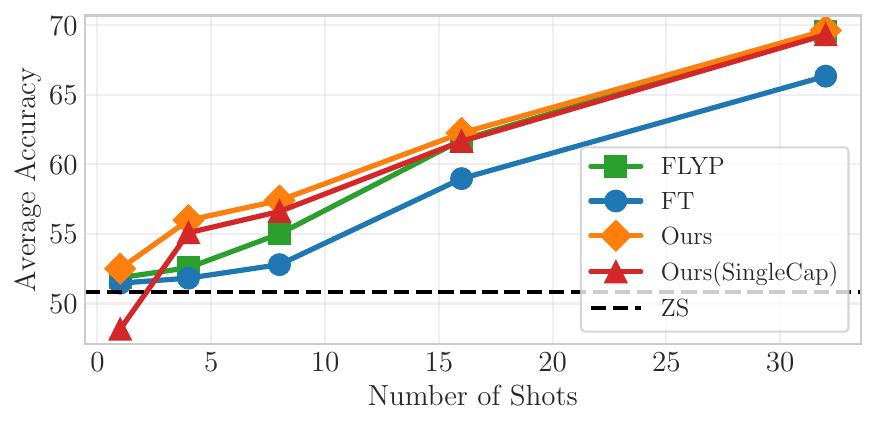}
\caption{
Average classification accuracy ($\%$) of ResNet-50 across 13 datasets with different numbers of training examples per class (shots).
Ours uses multiple captions focusing on three different characteristics: visual, shape, and texture, while Ours(SingleCap) uses a single caption focusing only on the visual characteristic.
}
\label{fig:few_shot}
\end{figure}

We evaluated our approach in few-shot learning scenarios using ResNet-50 as the backbone model across 13 datasets, including ImageNet.
For our method, we set the loss weight $w$ to 0.2 across all datasets.
Additionally, we evaluated a variant that uses only the ``visual'' characteristics within the prompt, demonstrating ``Ours(SingleCap)'', which generates a single caption per image, thereby proving the effectiveness of using multiple captions with different perspectives.
\Cref{fig:few_shot} shows the average classification accuracy across all datasets for different numbers of shots.
Detailed results for individual datasets are provided in the Appendix.

The results demonstrate that both Ours(SingleCap) and Ours significantly outperform baselines, particularly in low-shot regimes (K = 1, 4, 8).
At 4 shots, Ours achieves an average accuracy of $55.98\%$, outperforming FLYP ($52.57\%$) by 3.41 percentage points.
Moreover, Ours consistently outperforms Ours(SingleCap) when training data is extremely limited, though this gap narrows as the number of shots increases.
These results highlight the advantage of our approach in few-shot scenarios.
The captions generated by our method contain richer and more detailed information compared to the simple text templates used by baseline methods.
This allows our models to learn from more comprehensive information, even with limited training examples.
Ours, which uses three captions per image focusing on different visual aspects, provides more information for the model to learn from, explaining its superior performance in extremely low-shot settings.
As the number of shots increases to 16 and 32, the performance gap between different methods narrows.
This suggests that with sufficient training examples, even simpler approaches can learn adequate representations.
However, our method maintains a consistent advantage across all shot counts, demonstrating robustness and effectiveness in various data availability scenarios.

\subsection{Classification with Generated Captions without Model Training}

We evaluated the effectiveness of our caption generation approach in classification settings using ResNet-50 as the backbone model without any training.
\Cref{tab:caption_comparison} shows the classification results using different caption generation methods without any model training.

\begin{table*}[tb!]
\centering
\small
\caption{
Classification accuracy ($\%$) with ResNet-50 using different caption generation methods without model training. 
Comparative+Filtering uses different numbers of shots per dataset (ImageNet: 64, CIFAR100/CUB/Cars: 16, Food/Pet: 32, DTD: 8, Flowers: 4). 
Ours uses MLLM-generated captions with varying numbers of shots. 
For ImageNet with the full dataset, the result is not available due to the difficulty of generating captions for the entire training dataset.
Results for ViT-B/32 are in the Appendix.
}
\resizebox{\linewidth}{!}{
\begin{tabular}{lcccccccccc}
\toprule
Method & Shots & CIFAR100 & CUB & Cars & DTD & Flowers & Food & ImageNet & Pet & Average \\ \midrule
ZS & 0 & 40.98 & 42.13 & 50.99 & 39.04 & 56.25 & 76.67 & 56.98 & 80.57 & 55.45 \\
DCLIP & 0 &  43.18 & 43.77 & 53.04 & 43.46 & 63.59 & 78.06 & 59.67 & 81.71 & 58.31 \\
WaffleCLIP & 0 &  43.09 \fontsize{5pt}{5pt}\selectfont{$\pm$ 0.07} & 41.91 \fontsize{5pt}{5pt}\selectfont{$\pm$ 0.12} & 51.93 \fontsize{5pt}{5pt}\selectfont{$\pm$ 0.17} & 40.06 \fontsize{5pt}{5pt}\selectfont{$\pm$ 0.13} & 60.89 \fontsize{5pt}{5pt}\selectfont{$\pm$ 0.1} & 78.18 \fontsize{5pt}{5pt}\selectfont{$\pm$ 0.2} & 59.35 \fontsize{5pt}{5pt}\selectfont{$\pm$ 0.12} & 81.02 \fontsize{5pt}{5pt}\selectfont{$\pm$ 0.3} & 57.05 \\
Comparative & 0 &  43.87 & 43.61 & 52.62 & 43.35 & 64.03 & 79.13 & 59.82 & 83.02 & 58.68 \\ \midrule
Comparative + Filtering & $\leq 64$ &  43.65 \fontsize{5pt}{5pt}\selectfont{$\pm$ 0.11} & 42.73 \fontsize{5pt}{5pt}\selectfont{$\pm$ 0.08} & 49.71 \fontsize{5pt}{5pt}\selectfont{$\pm$ 0.35} & 49.37 \fontsize{5pt}{5pt}\selectfont{$\pm$ 0.7} & 65.12 \fontsize{5pt}{5pt}\selectfont{$\pm$ 0.51} & 79.40 \fontsize{5pt}{5pt}\selectfont{$\pm$ 0.09} & 60.47 \fontsize{5pt}{5pt}\selectfont{$\pm$ 0.06} & 80.41 \fontsize{5pt}{5pt}\selectfont{$\pm$ 0.12} & 58.86 \\
\rowcolor{lightgray}Ours w/o Training & 1 & 38.53 \fontsize{5pt}{5pt}\selectfont{$\pm$ 0.52} & 40.26 \fontsize{5pt}{5pt}\selectfont{$\pm$ 0.49} & 50.35 \fontsize{5pt}{5pt}\selectfont{$\pm$ 0.13} & 47.61 \fontsize{5pt}{5pt}\selectfont{$\pm$ 1.10} & 63.47 \fontsize{5pt}{5pt}\selectfont{$\pm$ 0.56} & 72.89 \fontsize{5pt}{5pt}\selectfont{$\pm$ 0.45} & 52.65 & 75.25 \fontsize{5pt}{5pt}\selectfont{$\pm$ 1.48} & 55.13 \\
\rowcolor{lightgray}Ours w/o Training & 4 & 41.83 \fontsize{5pt}{5pt}\selectfont{$\pm$ 0.31} & 42.79 \fontsize{5pt}{5pt}\selectfont{$\pm$ 0.09} & 54.22 \fontsize{5pt}{5pt}\selectfont{$\pm$ 0.21} & 55.27 \fontsize{5pt}{5pt}\selectfont{$\pm$ 0.31} & 66.16 \fontsize{5pt}{5pt}\selectfont{$\pm$ 0.19} & 78.50 \fontsize{5pt}{5pt}\selectfont{$\pm$ 0.08} & 58.08 & 82.33 \fontsize{5pt}{5pt}\selectfont{$\pm$ 0.35} & 59.90 \\
\rowcolor{lightgray}Ours w/o Training & 8 & 41.87 \fontsize{5pt}{5pt}\selectfont{$\pm$ 0.45} & 42.94 \fontsize{5pt}{5pt}\selectfont{$\pm$ 0.16} & 55.13 \fontsize{5pt}{5pt}\selectfont{$\pm$ 0.13} & 57.20 \fontsize{5pt}{5pt}\selectfont{$\pm$ 0.53} & 67.75 \fontsize{5pt}{5pt}\selectfont{$\pm$ 0.22} & 79.43 \fontsize{5pt}{5pt}\selectfont{$\pm$ 0.09} & 58.66 & 83.59 \fontsize{5pt}{5pt}\selectfont{$\pm$ 0.16} & 60.82 \\
\rowcolor{lightgray}Ours w/o Training & 16 & 41.95 \fontsize{5pt}{5pt}\selectfont{$\pm$ 0.26} & 43.61 \fontsize{5pt}{5pt}\selectfont{$\pm$ 0.19} & 55.96 \fontsize{5pt}{5pt}\selectfont{$\pm$ 0.07} & 58.65 \fontsize{5pt}{5pt}\selectfont{$\pm$ 0.09} & 67.88 \fontsize{5pt}{5pt}\selectfont{$\pm$ 0.00} & 79.83 \fontsize{5pt}{5pt}\selectfont{$\pm$ 0.09} & 59.46 & 84.40 \fontsize{5pt}{5pt}\selectfont{$\pm$ 0.48} & 61.47 \\
\rowcolor{lightgray}Ours w/o Training & 32 & 41.98 \fontsize{5pt}{5pt}\selectfont{$\pm$ 0.06} & 43.59 \fontsize{5pt}{5pt}\selectfont{$\pm$ 0.06} & 56.19 \fontsize{5pt}{5pt}\selectfont{$\pm$ 0.08} & 59.88 \fontsize{5pt}{5pt}\selectfont{$\pm$ 0.20} & 67.88 \fontsize{5pt}{5pt}\selectfont{$\pm$ 0.00} & 80.16 \fontsize{5pt}{5pt}\selectfont{$\pm$ 0.09} & 59.60 & 84.65 \fontsize{5pt}{5pt}\selectfont{$\pm$ 0.08} & 61.74 \\
\rowcolor{lightgray}Ours w/o Training & Full & 42.42 \fontsize{5pt}{5pt}\selectfont{$\pm$ 0.05} & 43.64 \fontsize{5pt}{5pt}\selectfont{$\pm$ 0.07} & 56.14 \fontsize{5pt}{5pt}\selectfont{$\pm$ 0.06} & 59.89 \fontsize{5pt}{5pt}\selectfont{$\pm$ 0.00} & 67.88 \fontsize{5pt}{5pt}\selectfont{$\pm$ 0.00} & 80.36 \fontsize{5pt}{5pt}\selectfont{$\pm$ 0.01} & - & 84.92 \fontsize{5pt}{5pt}\selectfont{$\pm$ 0.05} & - \\ \bottomrule

\end{tabular}
}

\label{tab:caption_comparison}
\end{table*}

With just 4 shots, our approach achieves an average accuracy of $59.90\%$, outperforming all baselines, including Comparative+Filtering ($58.86\%$), which uses substantially more shots for several datasets.
We observe varying performance across different dataset types.
For general object recognition datasets like CIFAR100 and ImageNet, our approach slightly underperforms compared to some baselines.
However, for fine-grained recognition datasets, our method demonstrates substantial improvements.
This is particularly evident in the DTD dataset, where our method with 32 shots achieves $59.88\%$ accuracy, outperforming the second-best baseline (Comparative+Filtering) by 10.51 percentage points.
These results highlight the ability of MLLM to generate highly informative captions that capture the discriminative features needed for effective classification, especially for fine-grained categories, even with limited examples.

\begin{figure}[t]
\centering
\includegraphics[width=0.9\linewidth]{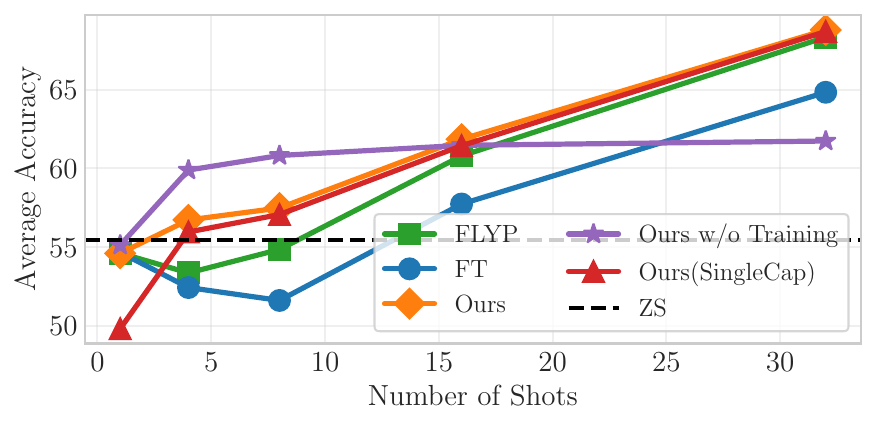}
\caption{
Comparison of average classification accuracy ($\%$) with ResNet-50 between fine-tuned approaches and caption-based approaches without training across 8 datasets. 
}
\label{fig:caption_vs_fst}
\end{figure}

To further analyze the effectiveness of our approach, we compare the performance of fine-tuned models and models using generated captions without training.
\Cref{fig:caption_vs_fst} shows the comparison of average accuracy across the same 8 datasets with different numbers of shots.

Our results reveal an interesting trend. 
In low-shot scenarios (1, 4, and 8 shots), our approach without training consistently outperforms all fine-tuned methods, including our own fine-tuned approach.
At 8 shots, our approach without training achieves $60.82\%$ accuracy, while Ours with fine-tuning reaches $57.49\%$, representing a 3.33 percentage point advantage for the no-training approach.
This performance gap can be attributed to the instability of fine-tuning with very limited data.
Fine-tuned models show higher variance in performance across datasets when training data is scarce, with some datasets experiencing significant performance drops.
In contrast, our approach without training demonstrates remarkable stability even with minimal examples.
As the number of shots increases to 16, both approaches achieve comparable performance.
However, with 32 shots, fine-tuned models begin to outperform the no-training approach.
These findings highlight an important practical consideration. 
When only a few labeled examples are available (8 or fewer shots per class), using our caption generation approach without any model training may be preferable to fine-tuning.
This can significantly reduce computational costs while achieving better performance in such resource-constrained scenarios.
Detailed comparison results for individual datasets are provided in the Appendix.

\subsection{Ablation Study on Caption Generation}

Our approach depends on the captions generated by MLLM, which can vary based on the prompt characteristics and the MLLM model used.
To investigate how these factors affect classification performance, we conducted an ablation study on three datasets (DTD, Flowers, and Pet) in the 8-shot learning scenario.
For these experiments, we use our fine-tuning method with the loss weight $w$ set to 0.2.
First, we investigate the impact of varying the prompt characteristics used in caption generation.
Additionally, we conducted experiments using text templates instead of captions and training with supervised contrastive fine-tuning loss.
\Cref{tab:ablation_captions} shows the classification accuracy across three datasets when using different captions.

\begin{table}[tb!]
\centering
\small
\caption{
Impact of different captions on classification accuracy (\%) in 8-shot learning scenario.
The upper section shows results with a single caption per image, while the lower section shows results with multiple captions per image.
}
\resizebox{\linewidth}{!}{
\begin{tabular}{lcccc}
\toprule
Caption & DTD & Flowers & Pet & Average \\ \midrule
Visual                       & 55.51 \fontsize{5pt}{5pt}\selectfont{$\pm$ 0.60} & 76.24 \fontsize{5pt}{5pt}\selectfont{$\pm$ 0.48} & 81.46 \fontsize{5pt}{5pt}\selectfont{$\pm$ 1.10} &  71.07       \\
Shape                                     & 54.06 \fontsize{5pt}{5pt}\selectfont{$\pm$ 0.37} & 76.18 \fontsize{5pt}{5pt}\selectfont{$\pm$ 0.16} & 81.19 \fontsize{5pt}{5pt}\selectfont{$\pm$ 0.32} &  70.48       \\
Texture                                  & 54.18 \fontsize{5pt}{5pt}\selectfont{$\pm$ 0.67} & 76.48 \fontsize{5pt}{5pt}\selectfont{$\pm$ 0.02} & 79.08 \fontsize{5pt}{5pt}\selectfont{$\pm$ 1.01} &   69.91      \\
Text template                                         & 47.46 \fontsize{5pt}{5pt}\selectfont{$\pm$ 0.50} & 76.66 \fontsize{5pt}{5pt}\selectfont{$\pm$ 0.03} & 77.91 \fontsize{5pt}{5pt}\selectfont{$\pm$ 1.37} & 67.34        \\
\midrule
Visual, Shape, Texture                   & 56.28 \fontsize{5pt}{5pt}\selectfont{$\pm$ 0.52} & 76.65 \fontsize{5pt}{5pt}\selectfont{$\pm$ 0.54} & 81.25 \fontsize{5pt}{5pt}\selectfont{$\pm$ 1.63} &  71.39       \\
\bottomrule
\end{tabular}
}
\label{tab:ablation_captions}
\end{table}

\begin{table}[tb!]
\centering
\small
\caption{
Impact of different MLLMs on classification accuracy ($\%$) in 8-shot learning scenario.
All MLLMs generate captions using the visual characteristic.
}
\resizebox{\linewidth}{!}{
\begin{tabular}{lcccc}
\toprule
MLLM                  & DTD                                       & Flowers                                   & Pet                                       & Average \\ \midrule
Gemini 2.5 Flash-Lite & 55.51 \fontsize{5pt}{5pt}\selectfont{$\pm$ 0.60} & 76.24 \fontsize{5pt}{5pt}\selectfont{$\pm$ 0.48} & 81.46 \fontsize{5pt}{5pt}\selectfont{$\pm$ 1.10} & 71.07   \\
Gemini 2.5 Flash      & 56.95 \fontsize{5pt}{5pt}\selectfont{$\pm$ 0.69} & 75.90 \fontsize{5pt}{5pt}\selectfont{$\pm$ 0.14} & 83.87 \fontsize{5pt}{5pt}\selectfont{$\pm$ 0.28} & 72.24   \\
Gemini 2.5 Pro        & 56.93 \fontsize{5pt}{5pt}\selectfont{$\pm$ 0.80} & 76.07 \fontsize{5pt}{5pt}\selectfont{$\pm$ 0.19} & 81.48 \fontsize{5pt}{5pt}\selectfont{$\pm$ 0.42} & 71.49   \\
GPT-4o mini           & 56.38 \fontsize{5pt}{5pt}\selectfont{$\pm$ 0.84} & 75.85 \fontsize{5pt}{5pt}\selectfont{$\pm$ 0.72} & 83.51 \fontsize{5pt}{5pt}\selectfont{$\pm$ 0.41} & 71.91   \\
GPT-5 mini            & 56.56 \fontsize{5pt}{5pt}\selectfont{$\pm$ 0.09} & 76.81 \fontsize{5pt}{5pt}\selectfont{$\pm$ 0.42} & 81.91 \fontsize{5pt}{5pt}\selectfont{$\pm$ 0.13} & 71.76   \\ \bottomrule
\end{tabular}
}
\label{tab:ablation_mllms}
\end{table}

When using a single caption per image, visual captions achieve the highest average accuracy, followed by shape and texture captions. 
Text templates show the lowest performance. The performance differences vary across datasets, with more pronounced gaps observed on DTD compared to Flowers and Pet. 
Combining multiple captions (visual, shape, texture) yields the best overall performance, though the improvement over single visual captions is modest.
Captions provide richer semantic information than simple text templates, enabling the model to learn more generalizable representations even with limited training examples.
The use of multiple diverse captions further enhances this generalization capability by capturing complementary aspects of visual content, allowing the model to build more robust feature representations.
However, the optimal caption type varies across datasets depending on their characteristics. 
For texture-focused datasets like DTD, captions that describe detailed visual patterns are more effective, while for object-centric datasets like Flowers and Pet, different caption types show similar effectiveness as the classification relies more on distinguishing well-defined categories. 
This dataset-dependent performance suggests that caption design should be tailored to the specific visual attributes relevant to each classification task.

Next, we examined the impact of different MLLMs.
\Cref{tab:ablation_mllms} compares the performance of five different MLLMs.: Gemini 2.5 Flash-Lite, Gemini 2.5 Flash, Gemini 2.5 Pro, GPT-4o mini~\citep{gpt4o}, and GPT-5 mini~\citep{openai2025gpt5}.

When comparing different MLLMs for generating visual captions, Gemini 2.5 Flash achieves the highest average accuracy, while Gemini 2.5 Flash-Lite shows the lowest.
There are performance differences across MLLMs, with approximately a 1 percentage point gap in average accuracy. 
The ranking of MLLMs varies by dataset: Gemini 2.5 Flash performs best on DTD and Pet, while GPT-5 mini achieves the highest accuracy on Flowers.
The dataset-dependent ranking suggests that different MLLMs may have distinct strengths in capturing certain types of visual features: some may excel at describing textures while others better articulate object-level characteristics. 

\subsection{Effect of Supervised Contrastive Loss Weight}

\begin{figure}[t]
\centering
\includegraphics[width=0.9\linewidth]{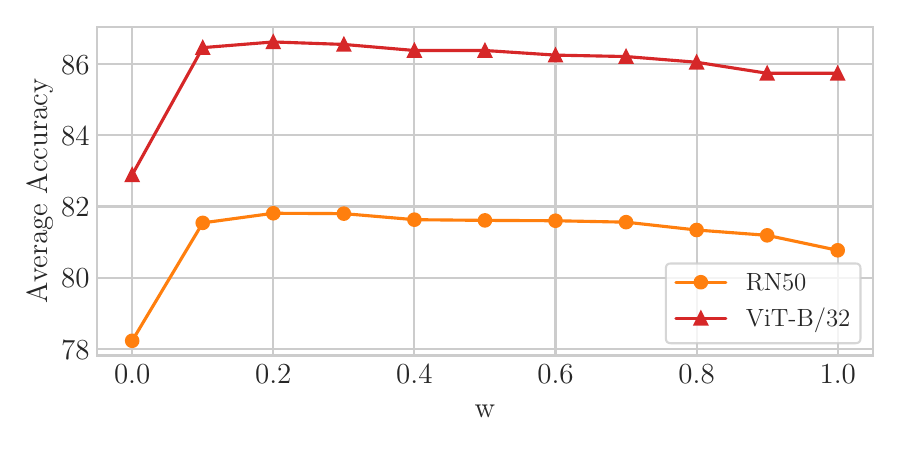}
\caption{Impact of supervised contrastive loss weight $w$ on average classification accuracy ($\%$) across 12 datasets (excluding ImageNet).}
\label{fig:hparams}
\end{figure}

We investigated the impact of the supervised contrastive loss weight $w$, which controls the balance between standard CLIP contrastive loss and our supervised contrastive loss. 
\Cref{fig:hparams} shows the average accuracy across 12 datasets (excluding ImageNet) for $w$ ranging from 0.0 to 1.0, for both ResNet-50 and ViT-B/32 backbones.

The results demonstrate that incorporating the supervised contrastive loss component significantly improves performance.
When $w=0$, which corresponds to using only the standard CLIP contrastive loss without class supervision, we observe the lowest accuracy for both backbones.
This performance drop occurs because the standard contrastive loss does not explicitly encourage embeddings from the same class to cluster together, potentially mapping images from the same class to different regions in the embedding space.
Both models show a similar trend where performance increases sharply as $w$ increases from 0 to 0.1, reaches a peak around $w=0.2$, and then gradually decreases as $w$ approaches 1.0.
This suggests that while class supervision is beneficial, it should be balanced with the original contrastive learning objective.
Interestingly, the performance remains relatively stable for $w$ values between 0.1 and 0.7, indicating that our approach is not overly sensitive to the exact choice of this hyperparameter.
This robustness to hyperparameter selection is advantageous for practical applications, as it reduces the need for extensive tuning when applying our method to new datasets.


\section{Limitations and Future Work}

While our approach demonstrates promising results, it is limited to image classification tasks in this paper.
Our study focuses on demonstrating the benefits of transforming unimodal datasets into multimodal ones as a proof of concept.
We showed that this data-centric approach provides significant benefits, particularly in few-shot scenarios.
The results indicate that MLLM-generated captions can effectively transfer the strong recognition capabilities of MLLMs to smaller models, even in extremely limited data settings, without requiring any model training.
However, many real-world applications require more complex tasks such as object detection and segmentation.
Extending our method to these tasks presents unique challenges, as MLLMs would need to generate not only descriptive captions but also spatial information about object locations and boundaries.
Additionally, our framework is currently limited to image datasets.
The core idea could be extended to other modalities, such as using audio-capable MLLMs to transform speech datasets into audio-text pairs.
These extensions to other tasks and modalities represent promising directions for future work.

\section{Conclusion}

In this study, we proposed a novel approach for enhancing image classification through multimodal fine-tuning. 
We demonstrated how MLLM can transform unimodal downstream task datasets into multimodal datasets by generating detailed and diverse synthetic captions for images. 
Our framework integrates a novel fine-tuning method specifically designed for synthetic multimodal datasets, combining supervised contrastive learning with standard CLIP contrastive loss to learn more discriminative representations. 
Additionally, we proposed a novel inference method that leverages information-rich synthetic captions through class-averaged text embedding inference, effectively using the semantic information captured in the generated captions to improve classification performance.

Our experimental results demonstrate consistent performance improvements over fine-tuning baselines across 13 image classification datasets. 
Particularly notable improvements were observed in few-shot learning scenarios, where our approach showed significant advantages. 
In the 8-shot scenario, our caption generation approach without any model training achieved $60.82\%$ accuracy, outperforming fine-tuned models by 3.33 percentage points while significantly reducing computational costs. 
The combination of multiple captions focusing on different visual characteristics (visual, shape, texture) further enhanced model performance, demonstrating the effectiveness of our multimodal approach in data-constrained environments.


{
    \small
    \bibliographystyle{ieeenat_fullname}
    \bibliography{main}
}


\clearpage
\appendix

\setcounter{table}{5}
\setcounter{figure}{4}
\setcounter{equation}{6}

\section{Details of the Main Experimental Results}

\subsection{Few-shot Learning Results Across All Datasets}

\begin{figure*}[tb!]
\centering
\includegraphics[width=1.0\linewidth]{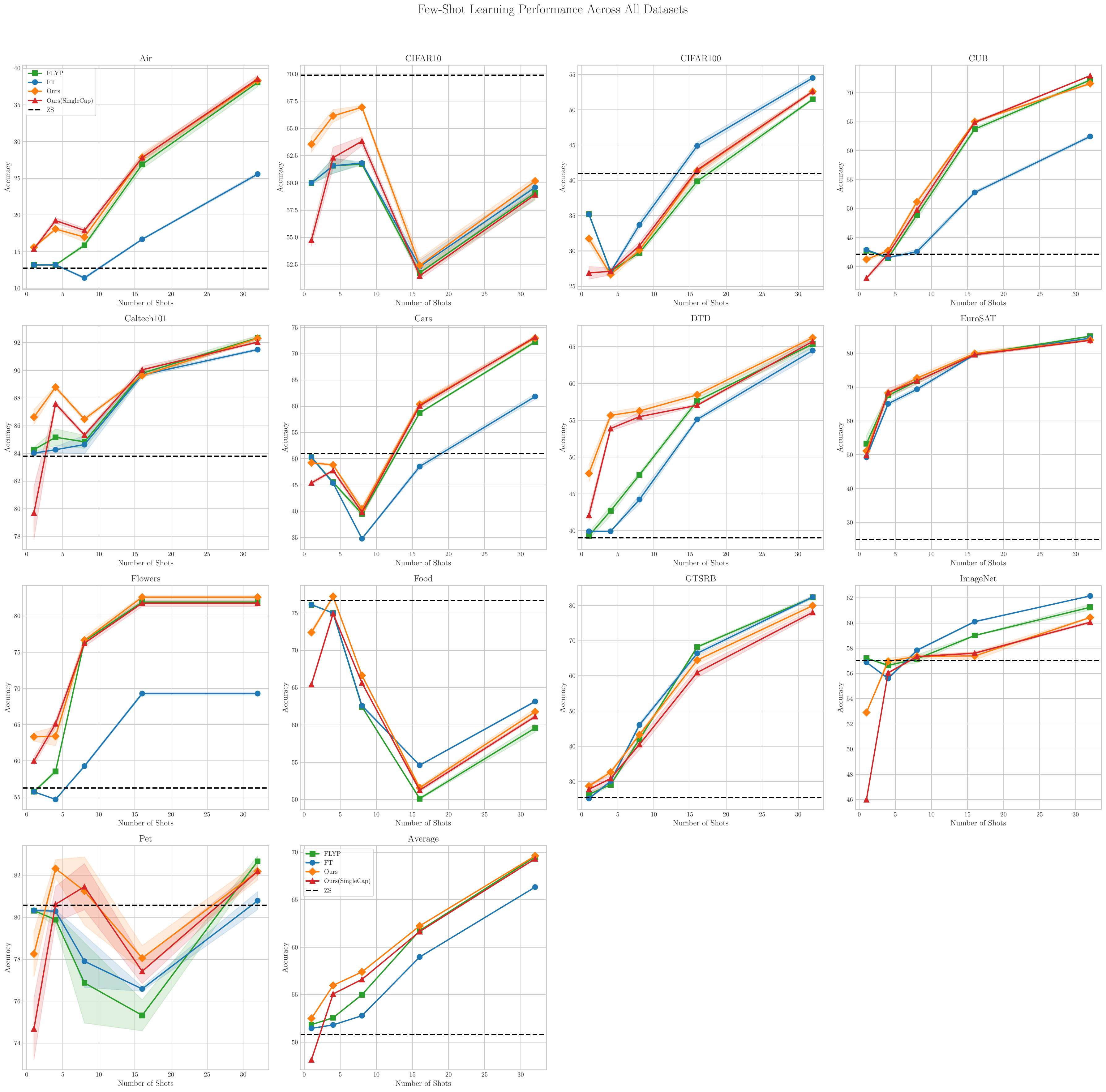}
\caption{
Results of few-shot learning for each dataset.
}
\label{fig:few_shot_all_datasets}
\end{figure*}

In the main paper, we reported the average accuracy across datasets; here, we provide the detailed results per dataset.
\Cref{fig:few_shot_all_datasets} shows the results.

\subsection{Detailed Results of Classification with Generated Captions without Model Training}

We evaluated the effectiveness of our caption generation approach in classification settings using ViT-B/32 as the backbone model without any training.
\Cref{tab:vit_caption} shows the classification results using different caption generation methods without any model training.

\begin{table*}[tb!]
\centering
\small
\caption{
Classification accuracy ($\%$) with ViT-B/32 using different caption generation methods without model training. 
Comparative+Filtering uses different numbers of shots per dataset (ImageNet/Food/Pet: 64, CIFAR100/Flowers: 32, CUB/DTD/Cars: 16). 
Ours uses MLLM-generated captions with varying numbers of shots. 
For ImageNet with the full dataset, the result is not available due to the difficulty of generating captions for the entire training dataset.
}
\resizebox{\linewidth}{!}{
\begin{tabular}{lcccccccccc}
\toprule
Method & Shots & CIFAR100 & CUB & Cars & DTD & Flowers & Food & ImageNet & Pet & Average \\ \midrule
ZS & 0 & 62.67 & 47.38 & 52.48 & 40.27 & 56.42 & 77.66 & 59.29 & 78.28 & 59.30 \\
DCLIP & 0 & 63.55 & 48.95 & 53.96 & 42.07 & 61.68 & 79.28 & 60.84 & 80.87 & 61.40 \\
WaffleCLIP & 0 & 63.67 \fontsize{5pt}{5pt}\selectfont{$\pm$ 0.07} & 47.76 \fontsize{5pt}{5pt}\selectfont{$\pm$ 0.11} & 53.11 \fontsize{5pt}{5pt}\selectfont{$\pm$ 0.1} & 38.99 \fontsize{5pt}{5pt}\selectfont{$\pm$ 0.22} & 57.44 \fontsize{5pt}{5pt}\selectfont{$\pm$ 0.12} & 78.95 \fontsize{5pt}{5pt}\selectfont{$\pm$ 0.19} & 60.05 \fontsize{5pt}{5pt}\selectfont{$\pm$ 0.05} & 80.37 \fontsize{5pt}{5pt}\selectfont{$\pm$ 0.23} & 60.04 \\
Comparative & 0 & 64.22 & 49.72 & 53.87 & 44.10 & 61.70 & 79.89 & 61.01 & 82.88 & 62.17 \\
\midrule
Comparative + Filtering & $\leq 64$ & 63.57 \fontsize{5pt}{5pt}\selectfont{$\pm$ 0.13} & 50.9 \fontsize{5pt}{5pt}\selectfont{$\pm$ 0.19} & 54.07 \fontsize{5pt}{5pt}\selectfont{$\pm$ 0.08} & 49.87 \fontsize{5pt}{5pt}\selectfont{$\pm$ 0.29} & 67.4 \fontsize{5pt}{5pt}\selectfont{$\pm$ 0.18} & 80.85 \fontsize{5pt}{5pt}\selectfont{$\pm$ 0.05} & 62.39 \fontsize{5pt}{5pt}\selectfont{$\pm$ 0.07} & 82.25 \fontsize{5pt}{5pt}\selectfont{$\pm$ 0.14} & 63.91 \\ 

\rowcolor{lightgray}Ours w/o Training & 1 & 58.54 \fontsize{5pt}{5pt}\selectfont{$\pm$ 0.18} & 49.11 \fontsize{5pt}{5pt}\selectfont{$\pm$ 0.20} & 51.87 \fontsize{5pt}{5pt}\selectfont{$\pm$ 0.12} & 47.39 \fontsize{5pt}{5pt}\selectfont{$\pm$ 0.52} & 67.84 \fontsize{5pt}{5pt}\selectfont{$\pm$ 0.77} & 75.41 \fontsize{5pt}{5pt}\selectfont{$\pm$ 0.42} & 56.42 \fontsize{5pt}{5pt}\selectfont{$\pm$ 0.00} & 84.86 \fontsize{5pt}{5pt}\selectfont{$\pm$ 0.20} & 61.43 \\ 

\rowcolor{lightgray}Ours w/o Training & 4 & 61.47 \fontsize{5pt}{5pt}\selectfont{$\pm$ 0.08} & 49.80 \fontsize{5pt}{5pt}\selectfont{$\pm$ 0.10} & 55.42 \fontsize{5pt}{5pt}\selectfont{$\pm$ 0.19} & 55.50 \fontsize{5pt}{5pt}\selectfont{$\pm$ 0.08} & 69.46 \fontsize{5pt}{5pt}\selectfont{$\pm$ 0.19} & 79.65 \fontsize{5pt}{5pt}\selectfont{$\pm$ 0.07} & 60.94 \fontsize{5pt}{5pt}\selectfont{$\pm$ 0.00} & 86.68 \fontsize{5pt}{5pt}\selectfont{$\pm$ 0.13} & 64.87  \\

\rowcolor{lightgray}Ours w/o Training & 8 & 61.61 \fontsize{5pt}{5pt}\selectfont{$\pm$ 0.13} & 49.93 \fontsize{5pt}{5pt}\selectfont{$\pm$ 0.11} & 55.77 \fontsize{5pt}{5pt}\selectfont{$\pm$ 0.18} & 56.97 \fontsize{5pt}{5pt}\selectfont{$\pm$ 0.54} & 69.40 \fontsize{5pt}{5pt}\selectfont{$\pm$ 0.16} & 80.27 \fontsize{5pt}{5pt}\selectfont{$\pm$ 0.10} & 61.53 \fontsize{5pt}{5pt}\selectfont{$\pm$ 0.00} & 87.03 \fontsize{5pt}{5pt}\selectfont{$\pm$ 0.24} & 65.31 \\ 

\rowcolor{lightgray}Ours w/o Training & 16 & 61.90 \fontsize{5pt}{5pt}\selectfont{$\pm$ 0.15} & 50.16 \fontsize{5pt}{5pt}\selectfont{$\pm$ 0.08} & 56.49 \fontsize{5pt}{5pt}\selectfont{$\pm$ 0.06} & 57.11 \fontsize{5pt}{5pt}\selectfont{$\pm$ 0.10} & 69.52 \fontsize{5pt}{5pt}\selectfont{$\pm$ 0.00} &  80.59 \fontsize{5pt}{5pt}\selectfont{$\pm$ 0.03} & 62.20 \fontsize{5pt}{5pt}\selectfont{$\pm$ 0.00} & 87.22 \fontsize{5pt}{5pt}\selectfont{$\pm$ 0.06} & 65.65 \\ 

\rowcolor{lightgray}Ours w/o Training & 32 & 62.30 \fontsize{5pt}{5pt}\selectfont{$\pm$ 0.06} & 50.05 \fontsize{5pt}{5pt}\selectfont{$\pm$ 0.07} & 56.85 \fontsize{5pt}{5pt}\selectfont{$\pm$ 0.06} & 58.03 \fontsize{5pt}{5pt}\selectfont{$\pm$ 0.11} & 69.52 \fontsize{5pt}{5pt}\selectfont{$\pm$ 0.00} & 80.84 \fontsize{5pt}{5pt}\selectfont{$\pm$ 0.07} & 62.31 \fontsize{5pt}{5pt}\selectfont{$\pm$ 0.00} & 87.63 \fontsize{5pt}{5pt}\selectfont{$\pm$ 0.07} & 65.94  \\ 

\rowcolor{lightgray}Ours w/o Training & Full & 62.79 \fontsize{5pt}{5pt}\selectfont{$\pm$ 0.04} & 50.03 \fontsize{5pt}{5pt}\selectfont{$\pm$ 0.01} & 56.78 \fontsize{5pt}{5pt}\selectfont{$\pm$ 0.02} & 58.24 \fontsize{5pt}{5pt}\selectfont{$\pm$ 0.00} & 69.52 \fontsize{5pt}{5pt}\selectfont{$\pm$ 0.00} & 80.99 \fontsize{5pt}{5pt}\selectfont{$\pm$ 0.02} & - & 87.70 \fontsize{5pt}{5pt}\selectfont{$\pm$ 0.03} & - \\ 
\bottomrule

\end{tabular}
}
\label{tab:vit_caption}
\end{table*}

The results for ViT-B/32 exhibit a similar trend to those of RN50, as described in the main text.
Our generated captions using 4-shot images demonstrate superior accuracy compared to all baselines.

\begin{figure*}[tb!]
\centering
\includegraphics[width=1.0\linewidth]{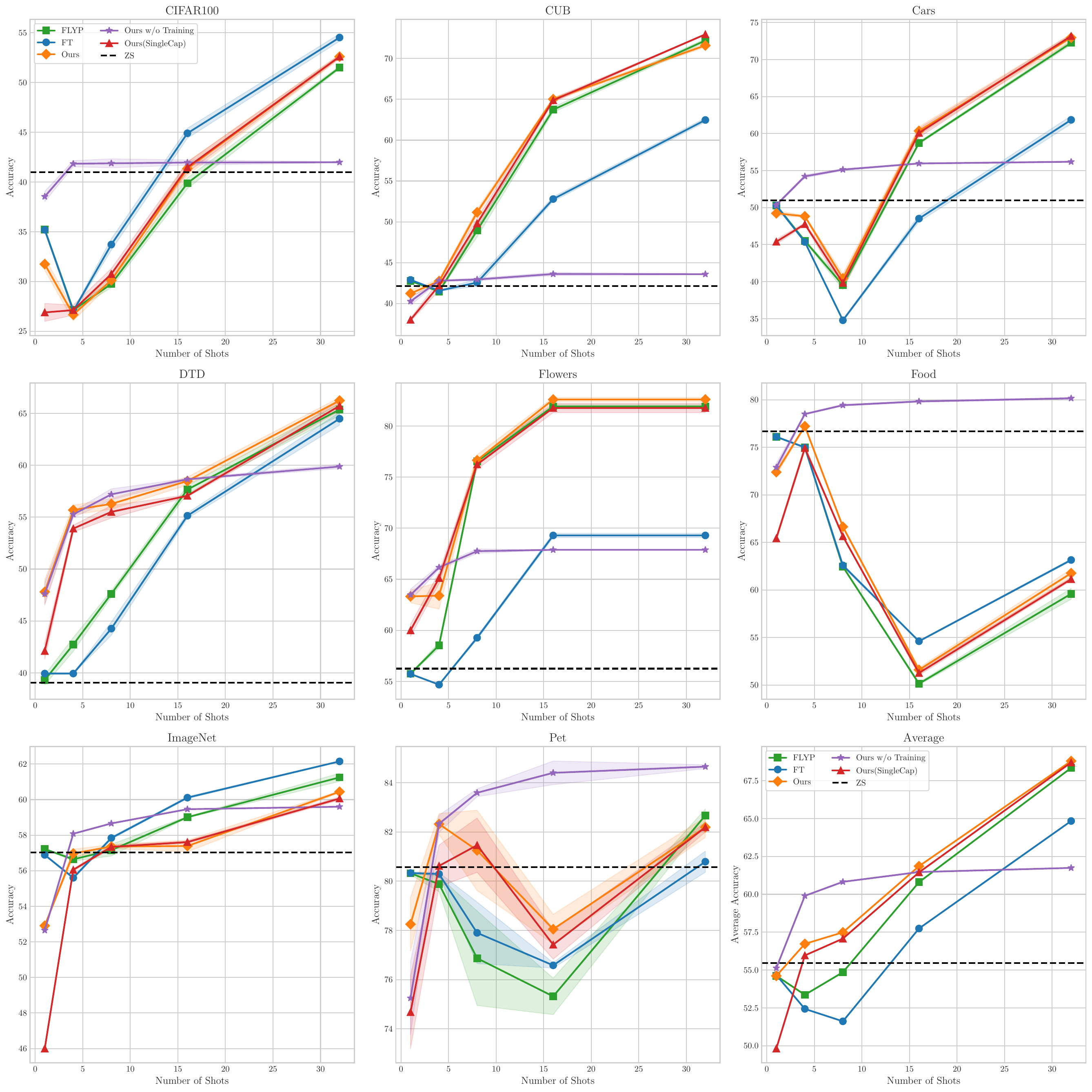}
\caption{
Comparison of classification accuracy ($\%$) with ResNet-50 between fine-tuned approaches and caption-based approaches without training for each dataset. 
}
\label{fig:caption_vs_fst_all_datasets}
\end{figure*}

Furthermore, \Cref{fig:caption_vs_fst_all_datasets} shows a comparison of the performance of the fine-tuned model and the model using captions generated without training across each dataset.

\subsection{Effect of Supervised Contrastive Loss Weight for Each Dataset}

\begin{figure*}[tb!]
\centering
\includegraphics[width=1.0\linewidth]{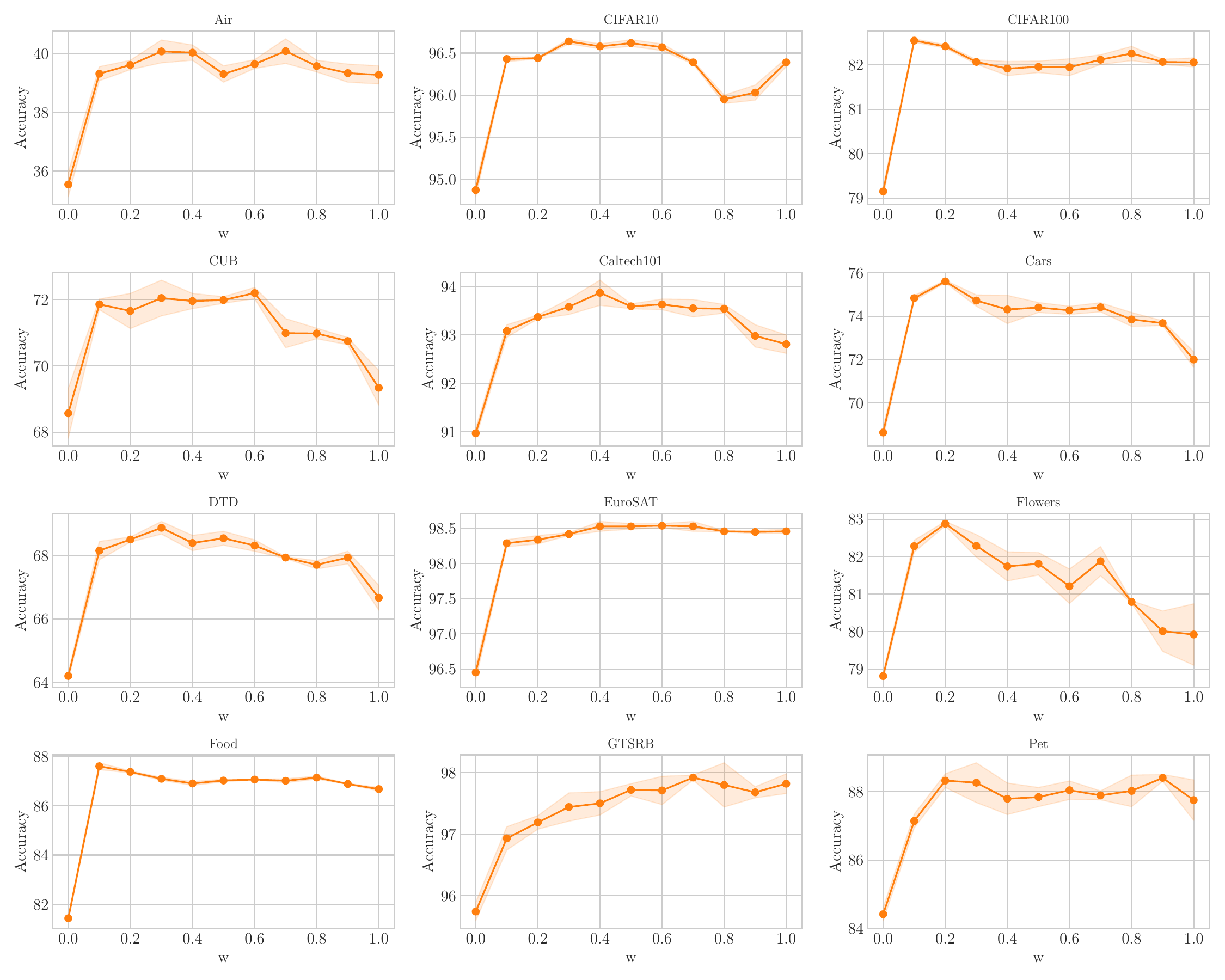}
\caption{
Classification accuracy for each dataset when varying the loss weight $w$ in the proposed method using RN50.
}
\label{fig:hparams_rn50}
\end{figure*}

\begin{figure*}[tb!]
\centering
\includegraphics[width=1.0\linewidth]{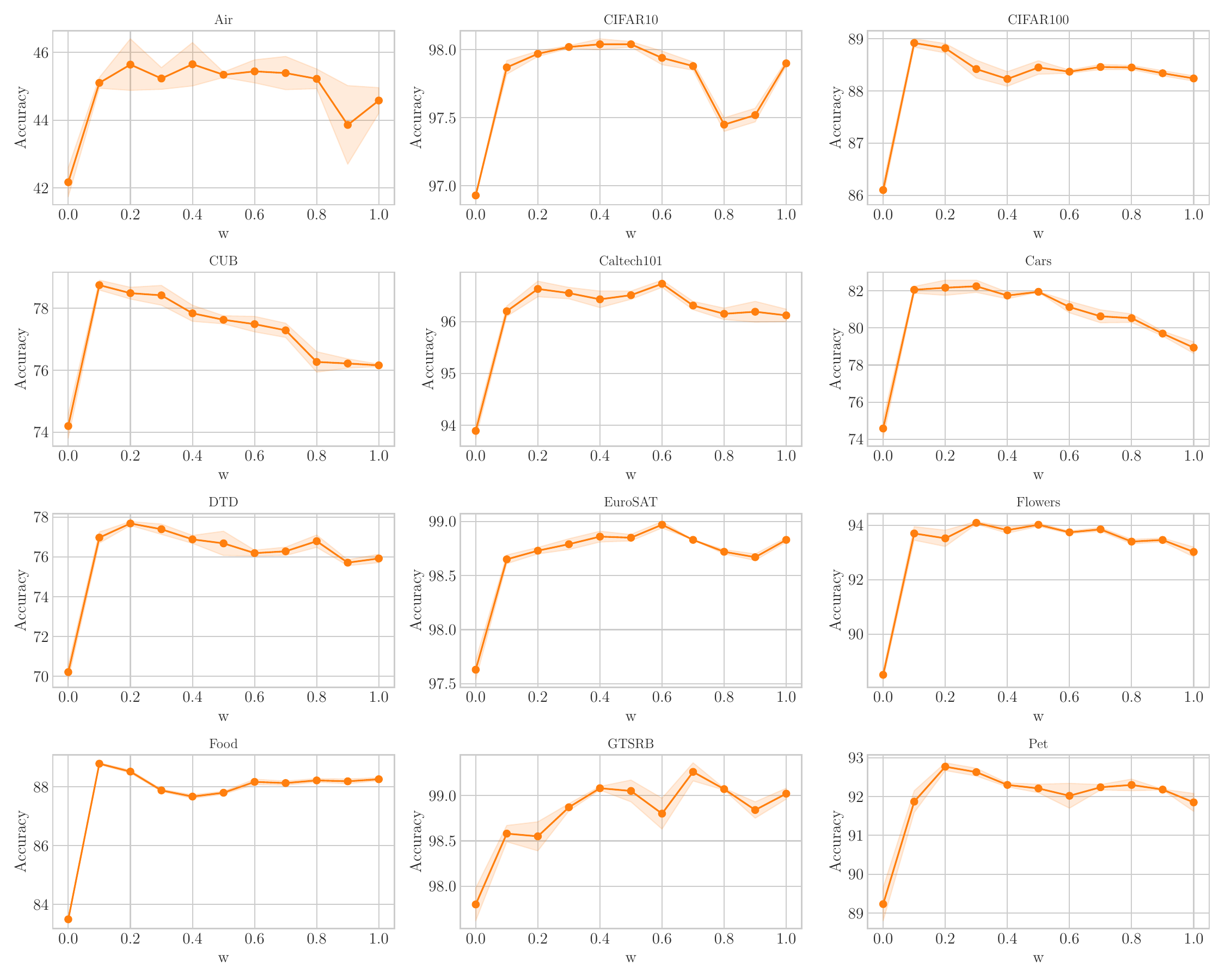}
\caption{
Classification accuracy for each dataset when varying the loss weight $w$ in the proposed method using ViT-B/32.
}
\label{fig:hparams_vit}
\end{figure*}

Here, we present the accuracy results for each dataset when varying the loss weight $w$ of the proposed method.
\Cref{fig:hparams_rn50} shows results for RN50, and \Cref{fig:hparams_vit} shows results for ViT-B/32.

The results indicate that the optimal $w$ varies for each dataset.

\section{Additional Experiments}

\subsection{Out-of-distribution Robustness}
We evaluated the robustness of the proposed method to out-of-distribution datasets using CIFAR10-C~\citep{in-c}, CIFAR10.1~\citep{recht2018cifar}, and CIFAR100-C~\citep{in-c}.
\Cref{tab:ood} shows the results.

Our method outperforms the baselines on all datasets.
This suggests that synthetic captions provide richer semantic representations that generalize better to distribution shifts.

\begin{table}[tb!]
\centering
\small
\caption{
Classification accuracy ($\%$) with RN50 on OOD testset. 
}
\label{tab:ood}
\begin{tabular}{lccc}
\toprule
Method & CIFAR10-C & CIFAR10.1 & CIFAR100-C \\ \midrule
ZS & 46.55 & 61.35 & 23.08 \\
FT & 73.97 \fontsize{5pt}{5pt}\selectfont{$\pm$ 0.72} & 89.62 \fontsize{5pt}{5pt}\selectfont{$\pm$ 0.65} & 51.96 \fontsize{5pt}{5pt}\selectfont{$\pm$ 0.19} \\
FLYP & 74.43 \fontsize{5pt}{5pt}\selectfont{$\pm$ 0.30} & 90.18 \fontsize{5pt}{5pt}\selectfont{$\pm$ 0.20} & 52.31 \fontsize{5pt}{5pt}\selectfont{$\pm$ 0.42} \\
\rowcolor{lightgray}Ours & \textbf{76.63} \fontsize{5pt}{5pt}\selectfont{$\pm$ 0.18} & \textbf{91.22} \fontsize{5pt}{5pt}\selectfont{$\pm$ 0.06} & \textbf{53.52} \fontsize{5pt}{5pt}\selectfont{$\pm$ 0.32} \\ \bottomrule
\end{tabular}
\end{table}

\subsection{Design Choice Analysis}

We conducted experiments on the design of the proposed method.
We conducted ablation experiments to validate key design choices in our framework: (1) adding text templates to generated captions, (2) the inference method for aggregating multiple caption embeddings, and (3) the choice between synthetic captions and text templates.
We evaluated five design variants against our default configuration:
\begin{itemize}
    \item \textbf{Default}: Our proposed method with class prefix concatenation and class-averaged text embedding inference
    \item \textbf{w/o Text Templates}: Generated captions without prepending text templates (``a photo of a \texttt{[class\_name]}'')
    \item \textbf{Logit-space Averaging Inference}: Class-wise averaging in logit space instead of embedding space
    \item \textbf{Nearest Caption Inference}: Nearest-neighbor inference using the most similar caption without averaging
    \item \textbf{Text Template Inference}: Using text templates instead of generated captions during inference
\end{itemize}
\Cref{tab:design} presents the results across 12 datasets using ResNet-50 backbone.

Removing text templates results in a slight performance drop, indicating that explicit class information helps ground the generated captions.
Alternative inference methods show comparable performance: embedding-space averaging performs similarly to logit-space averaging and nearest caption selection, suggesting robustness to the aggregation strategy.
Notably, embedding-space averaging is computationally more efficient as it computes similarities only with class prototypes (number of classes), whereas logit-space averaging and nearest caption selection require computing similarities with all individual captions (total number of captions), resulting in significantly higher computational cost.
However, replacing synthetic captions with text templates at inference time significantly degrades performance, confirming that our generated captions capture richer semantic information than simple templates.

\begin{table*}[tb!]
\centering
\small
\caption{
Ablation study on design choices with ResNet-50.
We evaluated five design variants across 12 datasets.
\textbf{Default}: our proposed method with class prefix concatenation and class-averaged text embedding inference.
\textbf{w/o Text Templates}: generated captions without prepending text templates.
\textbf{Logit-space Averaging Inference}: class-wise averaging in logit space instead of embedding space.
\textbf{Nearest Caption Inference}: nearest-neighbor inference without averaging.
\textbf{Text Template Inference}: using text templates instead of synthetic captions at inference.
}
\resizebox{\linewidth}{!}{
\begin{tabular}{lccccccccccccc}
\toprule
Design & Air & CIFAR10 & CIFAR100 & CUB & Caltech101 & Cars & DTD & EuroSAT & Flowers & Food & GTSRB & Pet & Average \\
\midrule
\rowcolor{lightgray}Default & 39.62 \fontsize{5pt}{5pt}\selectfont{$\pm$ 0.16} & 96.44 \fontsize{5pt}{5pt}\selectfont{$\pm$ 0.01} & 82.42 \fontsize{5pt}{5pt}\selectfont{$\pm$ 0.04} & 71.66 \fontsize{5pt}{5pt}\selectfont{$\pm$ 0.53} & 93.37 \fontsize{5pt}{5pt}\selectfont{$\pm$ 0.04} & 75.60 \fontsize{5pt}{5pt}\selectfont{$\pm$ 0.06} & 68.51 \fontsize{5pt}{5pt}\selectfont{$\pm$ 0.08} & 98.34 \fontsize{5pt}{5pt}\selectfont{$\pm$ 0.06} & 82.88 \fontsize{5pt}{5pt}\selectfont{$\pm$ 0.06} & 87.38 \fontsize{5pt}{5pt}\selectfont{$\pm$ 0.04} & 97.19 \fontsize{5pt}{5pt}\selectfont{$\pm$ 0.11} & 88.32 \fontsize{5pt}{5pt}\selectfont{$\pm$ 0.20} & 81.81 \\


w/o Text Templates & 38.42 \fontsize{5pt}{5pt}\selectfont{$\pm$ 0.11} & 96.53 \fontsize{5pt}{5pt}\selectfont{$\pm$ 0.05} & 82.21 \fontsize{5pt}{5pt}\selectfont{$\pm$ 0.18} & 70.64 \fontsize{5pt}{5pt}\selectfont{$\pm$ 0.19} & 93.70 \fontsize{5pt}{5pt}\selectfont{$\pm$ 0.20} & 74.85 \fontsize{5pt}{5pt}\selectfont{$\pm$ 0.19} & 69.77 \fontsize{5pt}{5pt}\selectfont{$\pm$ 0.47} & 98.25 \fontsize{5pt}{5pt}\selectfont{$\pm$ 0.03} & 81.41 \fontsize{5pt}{5pt}\selectfont{$\pm$ 0.31} & 87.41 \fontsize{5pt}{5pt}\selectfont{$\pm$ 0.09} & 97.37 \fontsize{5pt}{5pt}\selectfont{$\pm$ 0.15} & 88.15 \fontsize{5pt}{5pt}\selectfont{$\pm$ 0.05} & 81.56 \\

Logit-space Averaging Inference & 39.70 \fontsize{5pt}{5pt}\selectfont{$\pm$ 0.20} & 96.41 \fontsize{5pt}{5pt}\selectfont{$\pm$ 0.00} & 82.41 \fontsize{5pt}{5pt}\selectfont{$\pm$ 0.07} & 72.32 \fontsize{5pt}{5pt}\selectfont{$\pm$ 0.02} & 93.44 \fontsize{5pt}{5pt}\selectfont{$\pm$ 0.18} & 75.50 \fontsize{5pt}{5pt}\selectfont{$\pm$ 0.21} & 68.40 \fontsize{5pt}{5pt}\selectfont{$\pm$ 0.09} & 98.44 \fontsize{5pt}{5pt}\selectfont{$\pm$ 0.07} & 82.73 \fontsize{5pt}{5pt}\selectfont{$\pm$ 0.13} & 87.41 \fontsize{5pt}{5pt}\selectfont{$\pm$ 0.07} & 97.06 \fontsize{5pt}{5pt}\selectfont{$\pm$ 0.15} & 87.86 \fontsize{5pt}{5pt}\selectfont{$\pm$ 0.58} & 81.81 \\

Nearest Caption Inference & 39.92 \fontsize{5pt}{5pt}\selectfont{$\pm$ 0.27} & 96.53 \fontsize{5pt}{5pt}\selectfont{$\pm$ 0.06} & 82.39 \fontsize{5pt}{5pt}\selectfont{$\pm$ 0.09} & 72.40 \fontsize{5pt}{5pt}\selectfont{$\pm$ 0.27} & 93.50 \fontsize{5pt}{5pt}\selectfont{$\pm$ 0.05} & 75.36 \fontsize{5pt}{5pt}\selectfont{$\pm$ 0.15} & 68.24 \fontsize{5pt}{5pt}\selectfont{$\pm$ 0.18} & 98.43 \fontsize{5pt}{5pt}\selectfont{$\pm$ 0.00} & 83.09 \fontsize{5pt}{5pt}\selectfont{$\pm$ 0.09} & 87.39 \fontsize{5pt}{5pt}\selectfont{$\pm$ 0.02} & 97.45 \fontsize{5pt}{5pt}\selectfont{$\pm$ 0.23} & 87.51 \fontsize{5pt}{5pt}\selectfont{$\pm$ 0.78} & 81.85 \\

Text Template Inference & 38.17 \fontsize{5pt}{5pt}\selectfont{$\pm$ 0.01} & 96.43 \fontsize{5pt}{5pt}\selectfont{$\pm$ 0.05} & 82.26 \fontsize{5pt}{5pt}\selectfont{$\pm$ 0.06} & 71.39 \fontsize{5pt}{5pt}\selectfont{$\pm$ 0.42} & 93.51 \fontsize{5pt}{5pt}\selectfont{$\pm$ 0.10} & 75.14 \fontsize{5pt}{5pt}\selectfont{$\pm$ 0.39} & 65.35 \fontsize{5pt}{5pt}\selectfont{$\pm$ 0.11} & 98.34 \fontsize{5pt}{5pt}\selectfont{$\pm$ 0.04} & 78.23 \fontsize{5pt}{5pt}\selectfont{$\pm$ 0.08} & 87.29 \fontsize{5pt}{5pt}\selectfont{$\pm$ 0.02} & 97.44 \fontsize{5pt}{5pt}\selectfont{$\pm$ 0.03} & 87.07 \fontsize{5pt}{5pt}\selectfont{$\pm$ 0.46} & 80.89 \\

\bottomrule
\end{tabular}
}
\label{tab:design}
\end{table*}

\subsection{Effect of Batch Size}
Here, we show the results when varying the batch size for each fine-tuning method.
We conducted experiments with a learning rate of $1e-5$, 50 epochs, and batch sizes of 64, 128, 256, and 512.
While larger batch sizes are typically preferred for contrastive learning in pre-training~\citep{clip}, the optimal choice for fine-tuning downstream tasks remains unclear.
\Cref{tab:bs} shows the results.

Our method consistently outperforms baselines across all batch sizes, demonstrating robustness.
Notably, smaller batch sizes yield better performance: batch size 64 achieves the highest average accuracy (81.81\%), followed by 128 (80.15\%), 256 (78.63\%), and 512 (77.04\%).
This trend is consistent across all three methods (FT, FLYP, Ours), with performance degrading as batch size increases, though our method maintains its advantage over baselines at every setting.
The inverse relationship between batch size and fine-tuning performance contrasts with pre-training, where larger batches are preferred for contrastive learning. This difference arises because pre-training on diverse web-scale data benefits from large batches that capture varied negative pairs, whereas downstream datasets are more homogeneous with limited inherent diversity. In fine-tuning, smaller batches provide more frequent updates and stronger regularization, helping prevent overfitting to the task-specific distribution while preserving the generalization learned during pre-training.

\begin{table*}[tb!]
\centering
\small
\caption{
Impact of batch size on classification accuracy (\%) with ResNet-50 across 12 datasets.
We set $w=0.2$ for Ours.
}
\label{tab:bs}
\resizebox{\linewidth}{!}{
\begin{tabular}{llccccccccccccc}
\toprule
Batch Size           & Method & Air                                              & CIFAR10                                          & CIFAR100                                         & CUB                                              & Caltech101                                       & Cars                                             & DTD                                              & EuroSAT                                          & Flowers                                          & Food                                             & GTSRB                                            & Pet                                              & Average \\ \midrule
\multirow{3}{*}{64}  & FT     & 27.57 \fontsize{5pt}{5pt}\selectfont{$\pm$ 0.31} & 95.79 \fontsize{5pt}{5pt}\selectfont{$\pm$ 0.07} & 81.83 \fontsize{5pt}{5pt}\selectfont{$\pm$ 0.04} & 60.87 \fontsize{5pt}{5pt}\selectfont{$\pm$ 0.46} & 92.90 \fontsize{5pt}{5pt}\selectfont{$\pm$ 0.21} & 64.89 \fontsize{5pt}{5pt}\selectfont{$\pm$ 0.28} & 67.75 \fontsize{5pt}{5pt}\selectfont{$\pm$ 0.32} & 98.29 \fontsize{5pt}{5pt}\selectfont{$\pm$ 0.04} & 69.29 \fontsize{5pt}{5pt}\selectfont{$\pm$ 0.21} & 86.32 \fontsize{5pt}{5pt}\selectfont{$\pm$ 0.11} & 97.20 \fontsize{5pt}{5pt}\selectfont{$\pm$ 0.23} & 85.99 \fontsize{5pt}{5pt}\selectfont{$\pm$ 0.25} & 77.39   \\
                     & FLYP   & 38.91 \fontsize{5pt}{5pt}\selectfont{$\pm$ 0.59} & 95.93 \fontsize{5pt}{5pt}\selectfont{$\pm$ 0.07} & 82.18 \fontsize{5pt}{5pt}\selectfont{$\pm$ 0.20} & 71.52 \fontsize{5pt}{5pt}\selectfont{$\pm$ 0.22} & 93.41 \fontsize{5pt}{5pt}\selectfont{$\pm$ 0.10} & 74.92 \fontsize{5pt}{5pt}\selectfont{$\pm$ 0.43} & 68.05 \fontsize{5pt}{5pt}\selectfont{$\pm$ 0.54} & 98.47 \fontsize{5pt}{5pt}\selectfont{$\pm$ 0.03} & 81.92 \fontsize{5pt}{5pt}\selectfont{$\pm$ 0.20} & 86.93 \fontsize{5pt}{5pt}\selectfont{$\pm$ 0.02} & 97.77 \fontsize{5pt}{5pt}\selectfont{$\pm$ 0.07} & 87.83 \fontsize{5pt}{5pt}\selectfont{$\pm$ 0.32} & 81.49   \\
                 \rowcolor{lightgray}    & Ours   & 39.62 \fontsize{5pt}{5pt}\selectfont{$\pm$ 0.16} & 96.44 \fontsize{5pt}{5pt}\selectfont{$\pm$ 0.01} & 82.42 \fontsize{5pt}{5pt}\selectfont{$\pm$ 0.04} & 71.66 \fontsize{5pt}{5pt}\selectfont{$\pm$ 0.53} & 93.37 \fontsize{5pt}{5pt}\selectfont{$\pm$ 0.04} & 75.60 \fontsize{5pt}{5pt}\selectfont{$\pm$ 0.06} & 68.51 \fontsize{5pt}{5pt}\selectfont{$\pm$ 0.08} & 98.34 \fontsize{5pt}{5pt}\selectfont{$\pm$ 0.06} & 82.88 \fontsize{5pt}{5pt}\selectfont{$\pm$ 0.06} & 87.38 \fontsize{5pt}{5pt}\selectfont{$\pm$ 0.04} & 97.19 \fontsize{5pt}{5pt}\selectfont{$\pm$ 0.11} & 88.32 \fontsize{5pt}{5pt}\selectfont{$\pm$ 0.20} & 81.81   \\
                     \midrule
\multirow{3}{*}{128} & FT     & 23.61 \fontsize{5pt}{5pt}\selectfont{$\pm$ 0.21} & 95.43 \fontsize{5pt}{5pt}\selectfont{$\pm$ 0.12} & 80.31 \fontsize{5pt}{5pt}\selectfont{$\pm$ 0.17} & 58.20 \fontsize{5pt}{5pt}\selectfont{$\pm$ 0.23} & 91.78 \fontsize{5pt}{5pt}\selectfont{$\pm$ 0.07} & 60.43 \fontsize{5pt}{5pt}\selectfont{$\pm$ 0.27} & 64.96 \fontsize{5pt}{5pt}\selectfont{$\pm$ 0.11} & 98.12 \fontsize{5pt}{5pt}\selectfont{$\pm$ 0.05} & 63.72 \fontsize{5pt}{5pt}\selectfont{$\pm$ 0.43} & 85.86 \fontsize{5pt}{5pt}\selectfont{$\pm$ 0.05} & 96.49 \fontsize{5pt}{5pt}\selectfont{$\pm$ 0.32} & 85.23 \fontsize{5pt}{5pt}\selectfont{$\pm$ 0.28} & 75.35   \\
                     & FLYP   & 34.40 \fontsize{5pt}{5pt}\selectfont{$\pm$ 0.68} & 95.59 \fontsize{5pt}{5pt}\selectfont{$\pm$ 0.06} & 80.91 \fontsize{5pt}{5pt}\selectfont{$\pm$ 0.17} & 69.80 \fontsize{5pt}{5pt}\selectfont{$\pm$ 0.13} & 92.75 \fontsize{5pt}{5pt}\selectfont{$\pm$ 0.11} & 71.79 \fontsize{5pt}{5pt}\selectfont{$\pm$ 0.57} & 66.06 \fontsize{5pt}{5pt}\selectfont{$\pm$ 0.53} & 98.28 \fontsize{5pt}{5pt}\selectfont{$\pm$ 0.05} & 79.86 \fontsize{5pt}{5pt}\selectfont{$\pm$ 0.27} & 86.57 \fontsize{5pt}{5pt}\selectfont{$\pm$ 0.04} & 97.18 \fontsize{5pt}{5pt}\selectfont{$\pm$ 0.21} & 87.42 \fontsize{5pt}{5pt}\selectfont{$\pm$ 0.23} & 80.05   \\
                  \rowcolor{lightgray}   & Ours   & 34.49 \fontsize{5pt}{5pt}\selectfont{$\pm$ 0.07} & 95.86 \fontsize{5pt}{5pt}\selectfont{$\pm$ 0.05} & 81.65 \fontsize{5pt}{5pt}\selectfont{$\pm$ 0.19} & 69.85 \fontsize{5pt}{5pt}\selectfont{$\pm$ 0.14} & 92.59 \fontsize{5pt}{5pt}\selectfont{$\pm$ 0.16} & 71.87 \fontsize{5pt}{5pt}\selectfont{$\pm$ 0.39} & 67.11 \fontsize{5pt}{5pt}\selectfont{$\pm$ 0.43} & 98.14 \fontsize{5pt}{5pt}\selectfont{$\pm$ 0.02} & 80.14 \fontsize{5pt}{5pt}\selectfont{$\pm$ 0.35} & 86.98 \fontsize{5pt}{5pt}\selectfont{$\pm$ 0.12} & 96.42 \fontsize{5pt}{5pt}\selectfont{$\pm$ 0.14} & 86.71 \fontsize{5pt}{5pt}\selectfont{$\pm$ 0.01} & 80.15   \\
                     \midrule
\multirow{3}{*}{256} & FT     & 20.84 \fontsize{5pt}{5pt}\selectfont{$\pm$ 0.11} & 94.69 \fontsize{5pt}{5pt}\selectfont{$\pm$ 0.08} & 79.53 \fontsize{5pt}{5pt}\selectfont{$\pm$ 0.18} & 55.68 \fontsize{5pt}{5pt}\selectfont{$\pm$ 0.12} & 91.21 \fontsize{5pt}{5pt}\selectfont{$\pm$ 0.08} & 57.06 \fontsize{5pt}{5pt}\selectfont{$\pm$ 0.02} & 62.89 \fontsize{5pt}{5pt}\selectfont{$\pm$ 0.48} & 98.01 \fontsize{5pt}{5pt}\selectfont{$\pm$ 0.04} & 59.00 \fontsize{5pt}{5pt}\selectfont{$\pm$ 0.05} & 84.85 \fontsize{5pt}{5pt}\selectfont{$\pm$ 0.04} & 96.19 \fontsize{5pt}{5pt}\selectfont{$\pm$ 0.06} & 84.56 \fontsize{5pt}{5pt}\selectfont{$\pm$ 0.26} & 73.71   \\
                     & FLYP   & 30.05 \fontsize{5pt}{5pt}\selectfont{$\pm$ 0.08} & 95.07 \fontsize{5pt}{5pt}\selectfont{$\pm$ 0.09} & 80.25 \fontsize{5pt}{5pt}\selectfont{$\pm$ 0.08} & 66.41 \fontsize{5pt}{5pt}\selectfont{$\pm$ 0.13} & 91.89 \fontsize{5pt}{5pt}\selectfont{$\pm$ 0.18} & 68.94 \fontsize{5pt}{5pt}\selectfont{$\pm$ 0.14} & 65.04 \fontsize{5pt}{5pt}\selectfont{$\pm$ 0.17} & 98.05 \fontsize{5pt}{5pt}\selectfont{$\pm$ 0.06} & 77.61 \fontsize{5pt}{5pt}\selectfont{$\pm$ 0.14} & 85.55 \fontsize{5pt}{5pt}\selectfont{$\pm$ 0.17} & 96.67 \fontsize{5pt}{5pt}\selectfont{$\pm$ 0.12} & 86.08 \fontsize{5pt}{5pt}\selectfont{$\pm$ 0.23} & 78.47   \\
                 \rowcolor{lightgray}    & Ours   & 30.87 \fontsize{5pt}{5pt}\selectfont{$\pm$ 0.31} & 95.31 \fontsize{5pt}{5pt}\selectfont{$\pm$ 0.08} & 81.19 \fontsize{5pt}{5pt}\selectfont{$\pm$ 0.07} & 66.93 \fontsize{5pt}{5pt}\selectfont{$\pm$ 0.32} & 91.71 \fontsize{5pt}{5pt}\selectfont{$\pm$ 0.03} & 68.10 \fontsize{5pt}{5pt}\selectfont{$\pm$ 0.23} & 65.62 \fontsize{5pt}{5pt}\selectfont{$\pm$ 0.41} & 97.69 \fontsize{5pt}{5pt}\selectfont{$\pm$ 0.07} & 78.22 \fontsize{5pt}{5pt}\selectfont{$\pm$ 0.26} & 86.26 \fontsize{5pt}{5pt}\selectfont{$\pm$ 0.08} & 96.22 \fontsize{5pt}{5pt}\selectfont{$\pm$ 0.08} & 85.42 \fontsize{5pt}{5pt}\selectfont{$\pm$ 0.25} & 78.63   \\
\midrule
\multirow{3}{*}{512} & FT     & 17.19 \fontsize{5pt}{5pt}\selectfont{$\pm$ 0.13} & 93.92 \fontsize{5pt}{5pt}\selectfont{$\pm$ 0.08} & 77.79 \fontsize{5pt}{5pt}\selectfont{$\pm$ 0.19} & 52.99 \fontsize{5pt}{5pt}\selectfont{$\pm$ 0.22} & 90.18 \fontsize{5pt}{5pt}\selectfont{$\pm$ 0.22} & 53.89 \fontsize{5pt}{5pt}\selectfont{$\pm$ 0.16} & 61.03 \fontsize{5pt}{5pt}\selectfont{$\pm$ 0.18} & 97.49 \fontsize{5pt}{5pt}\selectfont{$\pm$ 0.02} & 52.34 \fontsize{5pt}{5pt}\selectfont{$\pm$ 0.69} & 84.00 \fontsize{5pt}{5pt}\selectfont{$\pm$ 0.09} & 95.29 \fontsize{5pt}{5pt}\selectfont{$\pm$ 0.11} & 84.23 \fontsize{5pt}{5pt}\selectfont{$\pm$ 0.09} & 71.70   \\
                     & FLYP   & 27.22 \fontsize{5pt}{5pt}\selectfont{$\pm$ 0.09} & 94.43 \fontsize{5pt}{5pt}\selectfont{$\pm$ 0.10} & 78.87 \fontsize{5pt}{5pt}\selectfont{$\pm$ 0.25} & 64.57 \fontsize{5pt}{5pt}\selectfont{$\pm$ 0.22} & 91.49 \fontsize{5pt}{5pt}\selectfont{$\pm$ 0.27} & 64.41 \fontsize{5pt}{5pt}\selectfont{$\pm$ 0.14} & 62.71 \fontsize{5pt}{5pt}\selectfont{$\pm$ 0.37} & 97.68 \fontsize{5pt}{5pt}\selectfont{$\pm$ 0.07} & 73.73 \fontsize{5pt}{5pt}\selectfont{$\pm$ 0.08} & 84.39 \fontsize{5pt}{5pt}\selectfont{$\pm$ 0.08} & 96.13 \fontsize{5pt}{5pt}\selectfont{$\pm$ 0.08} & 85.46 \fontsize{5pt}{5pt}\selectfont{$\pm$ 0.09} & 76.76   \\
                  \rowcolor{lightgray}   & Ours   & 28.82 \fontsize{5pt}{5pt}\selectfont{$\pm$ 0.16} & 94.91 \fontsize{5pt}{5pt}\selectfont{$\pm$ 0.02} & 79.52 \fontsize{5pt}{5pt}\selectfont{$\pm$ 0.11} & 65.43 \fontsize{5pt}{5pt}\selectfont{$\pm$ 0.18} & 90.55 \fontsize{5pt}{5pt}\selectfont{$\pm$ 0.30} & 63.69 \fontsize{5pt}{5pt}\selectfont{$\pm$ 0.25} & 64.36 \fontsize{5pt}{5pt}\selectfont{$\pm$ 0.21} & 97.19 \fontsize{5pt}{5pt}\selectfont{$\pm$ 0.04} & 75.14 \fontsize{5pt}{5pt}\selectfont{$\pm$ 0.23} & 85.40 \fontsize{5pt}{5pt}\selectfont{$\pm$ 0.03} & 94.63 \fontsize{5pt}{5pt}\selectfont{$\pm$ 0.14} & 84.87 \fontsize{5pt}{5pt}\selectfont{$\pm$ 0.03} & 77.04   \\ 
\bottomrule
\end{tabular}
}
\end{table*}

\subsection{Caption Examples}

\begin{figure*}[tb!]
\centering
\small
\setlength{\tabcolsep}{4pt}
\begin{tabular}{c|p{4.5cm}|p{4.5cm}|p{4.5cm}}
\toprule
\textbf{Image} & \textbf{Visual Caption} & \textbf{Shape Caption} & \textbf{Texture Caption} \\
\midrule
\begin{tabular}{@{}c@{}}
\resizebox{0.15\textwidth}{!}{\includegraphics{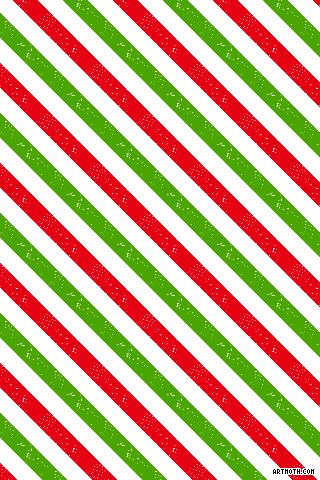}} \\
\textbf{DTD} \\
\textit{Class: Banded}
\end{tabular}
& 
This image features a repeating pattern of diagonal stripes in alternating red and green. The stripes have a subtle, grainy texture, giving them a slightly distressed or frosted appearance. The consistent spacing and angle of the bands create a dynamic, yet orderly visual rhythm.
& 
The primary shape characteristic is a series of parallel, diagonal bands. These bands alternate in color, red and green, and appear to be of uniform width. The overall impression is a repeating, linear pattern with a consistent angle.
& 
This texture features bold, diagonal stripes of red and green alternating with white. The colored stripes have a subtle, speckled or grainy quality, adding a touch of visual noise that differentiates them from a flat, uniform color. The overall impression is one of crisp, repeating patterns.
\\
\midrule
\begin{tabular}{@{}c@{}}
\resizebox{0.15\textwidth}{!}{\includegraphics{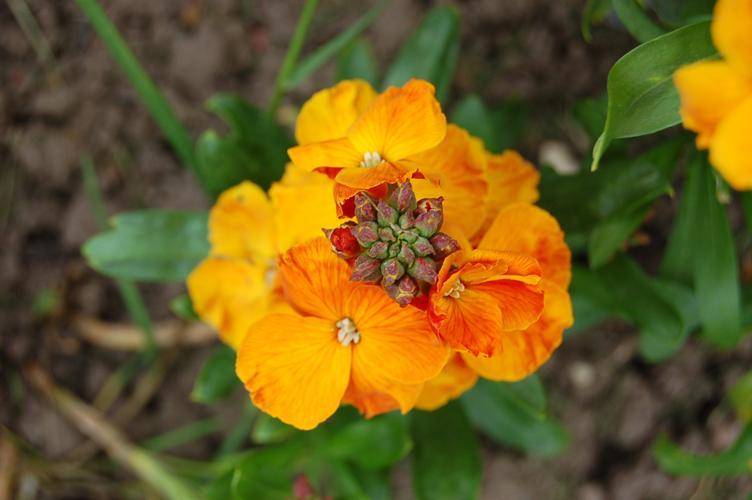}} \\
\textbf{Flowers} \\
\textit{Class: Wallflower}
\end{tabular}
& 
This wallflower features vibrant, clustered orange and yellow petals with a central cluster of unopened buds. Its rich, warm colors and dense bloom differentiate it, set against a soft, earthy background of green leaves and soil.
& 
The wallflower exhibits a cluster of vibrant, orange-yellow flowers. Each flower has four rounded petals, forming a somewhat star-like or cross shape. A central cluster of unopened buds adds a textured, conical element to the overall composition.
& 
The wallflower's petals appear velvety and soft, with a slightly matte finish. The central buds offer a contrasting texture, appearing more rigid and bumpy, like tiny, tightly packed beads. The overall impression is one of delicate, layered softness.
\\
\midrule
\begin{tabular}{@{}c@{}}
\resizebox{0.15\textwidth}{!}{\includegraphics{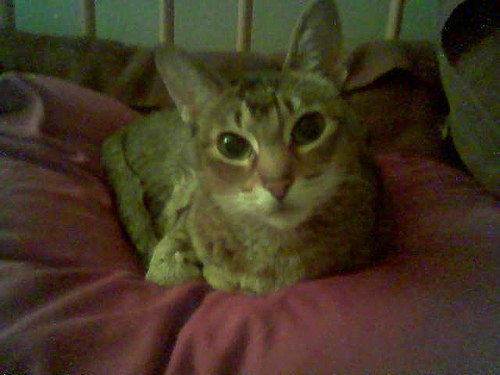}} \\
\textbf{Pet} \\
\textit{Class: Abyssinian}
\end{tabular}
& 
This Abyssinian cat features large, expressive eyes, prominent ears, and a ticked coat pattern. Its slender build and alert posture suggest an active and curious nature, setting it apart from more common feline appearances.
& 
This Abyssinian cat exhibits a lean, athletic build. Its body is elongated and slender, with a narrow chest and a tucked abdomen. The legs are long and graceful, contributing to its overall lithe silhouette. The head is wedge-shaped with large, expressive eyes and prominent ears.
& 
The Abyssinian cat's fur appears short and dense, with a subtle ticked pattern that creates a soft, almost velvety texture. The individual hairs seem fine, contributing to a sleek and smooth overall feel.
\\
\bottomrule
\end{tabular}
\caption{
Examples of synthetic captions generated for three different datasets (DTD, Flowers, and Pet). 
For each dataset, we show an example image from one class and the corresponding visual, shape, and texture captions generated by Gemini 2.5 Flash-lite. 
The captions demonstrate how different prompt types capture complementary aspects of the visual content, providing rich semantic information for fine-tuning.
}
\label{fig:caption_examples}
\end{figure*}

Here, we provide examples of captions generated by MLLM.
We selected one image each from the DTD, Flowers, and Pet datasets and show the corresponding visual, shape, and texture captions for each image.
\Cref{fig:caption_examples} shows examples of images and their corresponding captions.

\subsection{Analysis of Generated Captions}

We analyze the quality and characteristics of generated captions to validate our caption generation strategy.
Specifically, we examined CLIP scores (measuring image-text alignment) and token lengths for captions generated on the Pet dataset.
All generated captions have the text template (``a photo of a [class\_name]'') prepended.

\begin{figure}[tb!]
\centering
\includegraphics[width=1.0\linewidth]{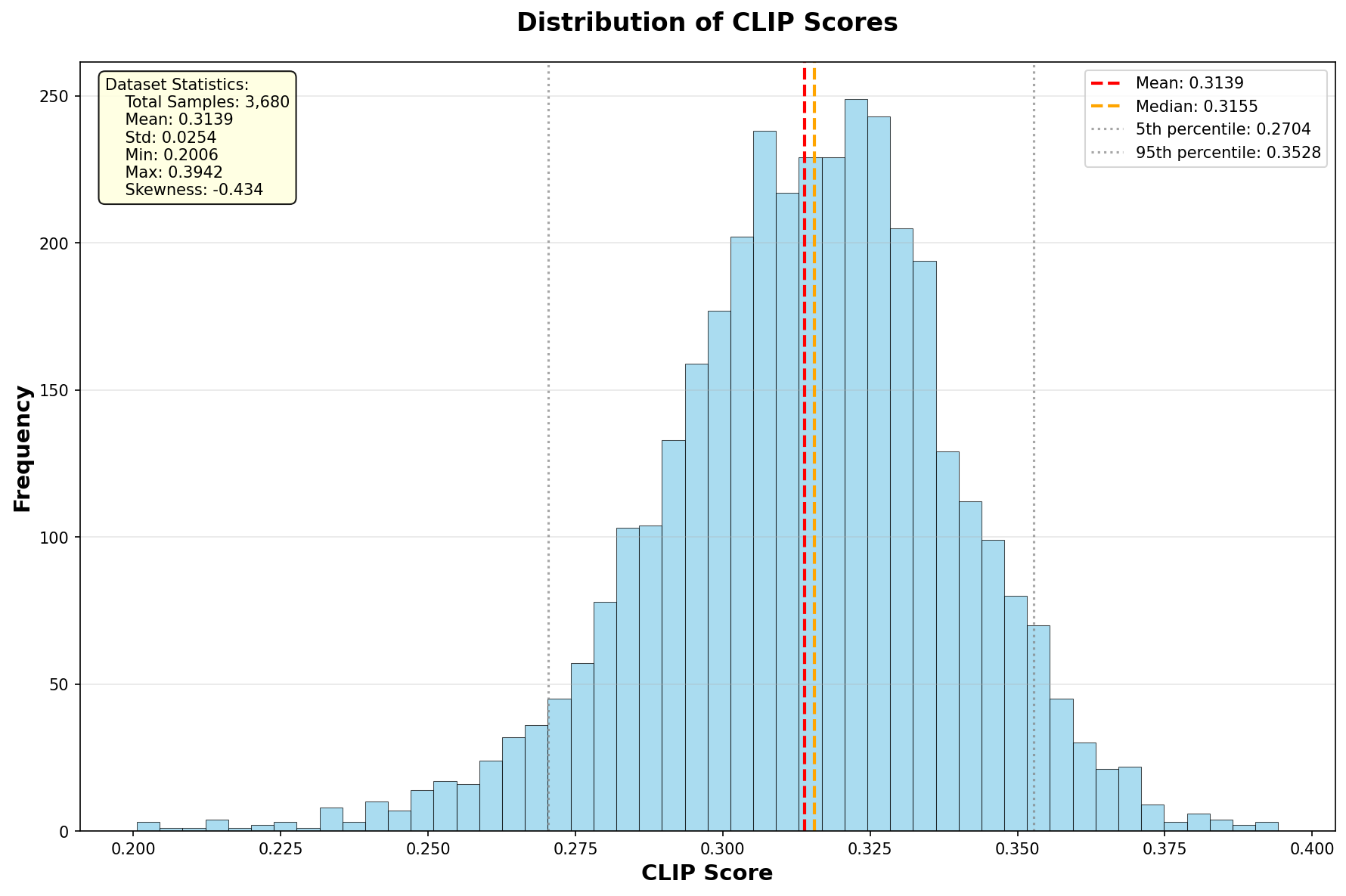}
\caption{
Distribution of CLIP scores for generated captions on the Pet dataset.
CLIP score is computed as the cosine similarity between image and text embeddings in CLIP's feature space.
Higher scores indicate stronger semantic correspondence between images and their synthetic captions.
}
\label{fig:clip_score}
\end{figure}

\Cref{fig:clip_score} shows the distribution of CLIP scores, which measure the cosine similarity between image and text embeddings in CLIP's feature space.
The histogram reveals that most generated captions achieve high CLIP scores (mean: 0.3139), indicating strong semantic alignment between the synthetic text and corresponding images.
This high alignment suggests that MLLM successfully captures visual content in textual form, providing meaningful supervision signals for fine-tuning.

\begin{figure}[tb!]
\centering
\includegraphics[width=1.0\linewidth]{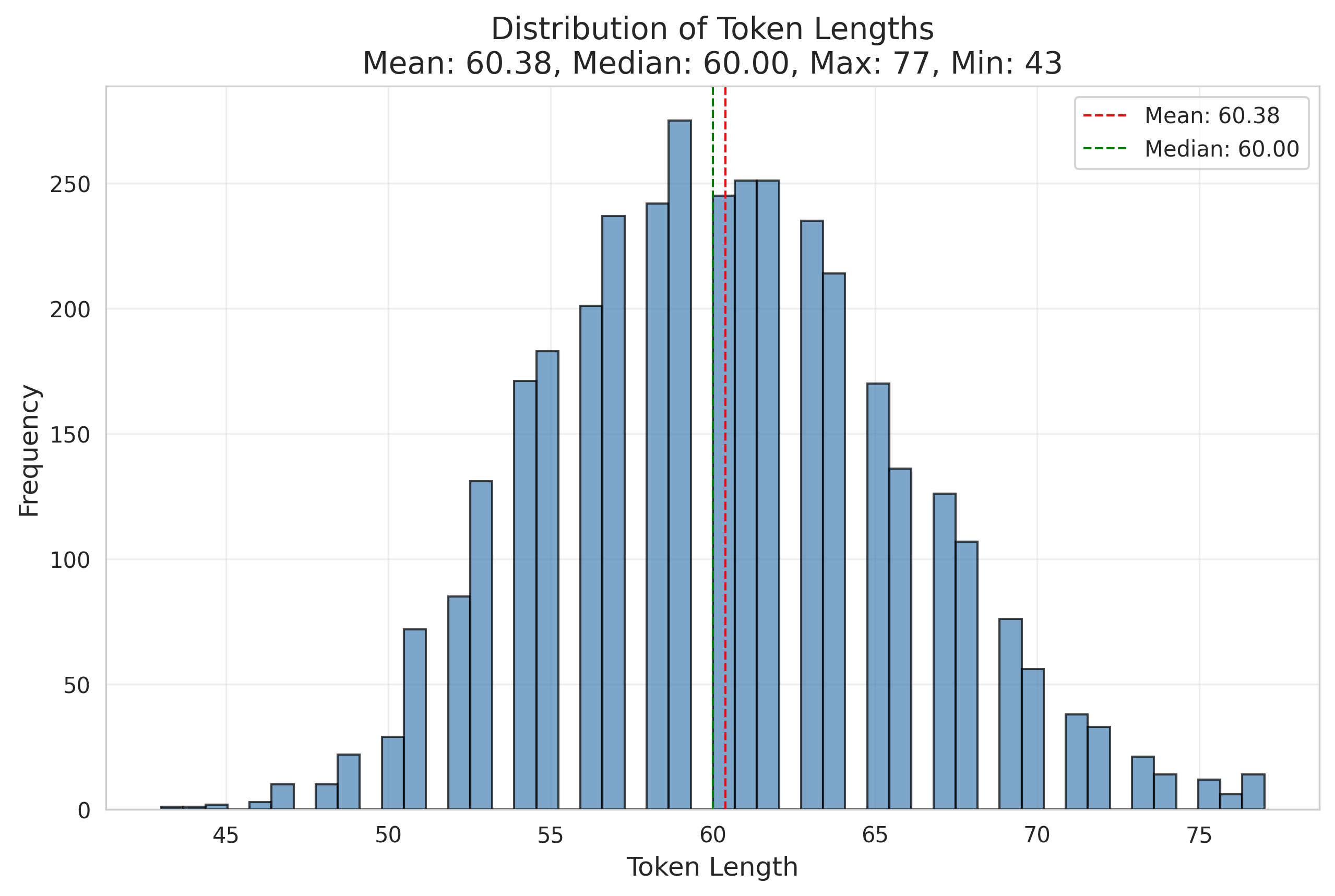}
\caption{
Distribution of token lengths for generated captions on the Pet dataset.
Token length is measured after tokenization.
}
\label{fig:token_length}
\end{figure}

\Cref{fig:token_length} shows the distribution of token lengths after tokenization with CLIP's text encoder.
Our approach specifies caption lengths of approximately 50 words in the generation prompts, which, combined with the prepended text template, results in token lengths well within CLIP's maximum context length of 77 tokens (mean: 60.38 tokens).
This design choice ensures that all captions can be fully processed without truncation, preserving the complete semantic information in each generated description.
The relatively consistent token lengths across samples also indicate that our prompting strategy successfully controls caption verbosity, avoiding both overly brief descriptions and excessively long texts.

\end{document}